\documentclass[lettersize,journal]{IEEEtran}
\usepackage{amsmath,amsfonts,amssymb}   
\usepackage{algorithm}                  
\usepackage{algorithmic}                
\usepackage{array}                      
\usepackage[caption=false,font=normalsize,labelfont=sf,textfont=sf]{subfig}  
\usepackage{textcomp}                   
\usepackage{stfloats}                   
\usepackage{url}                        
\usepackage{verbatim}                   
\usepackage{graphicx}                   
\usepackage{cite}                       
\usepackage{color}                      
\usepackage{xcolor}                     
\usepackage{multirow}                   
\usepackage{pifont}                     
\usepackage{makecell}                   
\usepackage[normalem]{ulem}             
\usepackage{booktabs}                   
\usepackage{tcolorbox}                  
\usepackage{placeins}                   
\usepackage{colortbl}                   
\usepackage{threeparttable}             
\usepackage{tabularx}                   
\usepackage{wrapfig}                    
\usepackage{microtype}                  
\usepackage{newfloat}                   
\usepackage{listings}                   
\usepackage{dcolumn}                    
\usepackage{subcaption}                 
\usepackage{svg}                        

\definecolor{darkgreen}{rgb}{0.0, 0.5, 0.0}
\definecolor{darkred}{rgb}{0.7, 0.0, 0.2}
\definecolor{darkblue}{rgb}{0,  .3,  .7}

\usepackage{tikz, xcolor}

\definecolor{lime}{HTML}{A6CE39}
\DeclareRobustCommand{\orcidicon}{
    \begin{tikzpicture}
        \draw[lime, fill=lime] (0,0)
        circle[radius=0.13]
        node[white]{{\fontfamily{qag}\selectfont \tiny \.{I}D}};
    \end{tikzpicture}
    \hspace{-2mm}
}
\foreach \x in {A, ..., Z}{%
    \expandafter\xdef\csname orcid\x\endcsname{\noexpand\href{https://orcid.org/\csname orcidauthor\x\endcsname}{\noexpand\orcidicon}}
}
\usepackage[implicit=false]{hyperref}

\hypersetup{
    hidelinks,         
    colorlinks=true,   
    allcolors=black,   
    pdfstartview=Fit,  
    breaklinks=true,    
    implicit=false
}
 
\begin{document}

\title{AttriReBoost: A Gradient-Free Propagation Optimization Method for Cold Start Mitigation in Attribute Missing Graphs}

\author{
Mengran~Li\hspace{-2mm}\orcidA{}\hspace{-1mm},
Chaojun~Ding\hspace{-2mm}\orcidD{}\hspace{-1mm},
Junzhou~Chen\hspace{-2mm}\orcidC{}\hspace{-1mm},
Wenbin Xing\hspace{-2mm}\orcidF{}\hspace{-1mm},
Cong Ye\hspace{-2mm}\orcidE{}\hspace{-1mm},
Ronghui~Zhang\hspace{-2mm}\orcidB{}\hspace{-1mm},\\
Songlin~Zhuang\hspace{-2mm}\orcidG{}\hspace{-1mm}, 
Jia Hu\hspace{-2mm}\orcidH{}\hspace{-1mm}, 
Tony Z. Qiu\hspace{-2mm}\orcidJ{}\hspace{-1mm}, and
Huijun Gao\hspace{-2mm}\orcidI{}\hspace{-1mm},~\IEEEmembership{Fellow,~IEEE}

\IEEEcompsocitemizethanks{

\IEEEcompsocthanksitem This work has been submitted to the lEEE for possible publicationCopyright may be transferred without notice, after which this version mayno longer be accessible.
\IEEEcompsocthanksitem This project is jointly supported by the Shenzhen Fundamental Research Program (No. JCYJ20240813151129038), the National Natural Science Foundation of China (Nos. 52172350, 51775565), the Guangdong Basic and Applied Research Foundation (No. 2022B1515120072), the Guangzhou Science and Technology Plan Project (No. 2024B01W0079), the Nansha Key R\&D Program (No. 2022ZD014), and the China Postdoctoral Science Foundation (No. 2013T60904). \textit{(Corresponding author: Ronghui Zhang.)}
\IEEEcompsocthanksitem  Mengran~Li, Junzhou Chen, Wenbin Xing, Ye Cong, and Ronghui~Zhang are with School of Intelligent Systems Engineering, Shenzhen Campus of Sun Yat-sen University, Shenzhen 518107, P.R. China. Ronghui Zhang is also with the School of Transportation Science and Engineering, Harbin Institute of Technology, Harbin 150080, P.R. China (e-mail: limr39@mail2.sysu.edu.cn; chenjunzhou@mail.sysu.edu.cn; {xingwb, yecong5}@mail2.sysu.edu.cn; zhangrh25@mail.sysu.edu.cn).
\IEEEcompsocthanksitem  
Songlin Zhuang is with the Yongjiang Laboratory, Ningbo 315202, P.R. China (e-mail: songlin-zhuang@ylab.ac.cn).
\IEEEcompsocthanksitem  
Jia Hu is with the Key Laboratory of Road and Traffic Engineering of the Ministry of Education, Tongji University, Shanghai 201804, P.R. China (e-mail: hujia@tongji.edu.cn). 
\IEEEcompsocthanksitem 
Tony Z. Qiu is with Department of Civil and Environmental Engineering, University of Alberta, Edmonton, AB T6G 2R3, Canada (e-mail: zhijunqiu@ualberta.ca).
\IEEEcompsocthanksitem  
Huijun Gao is with the Research Institute of Intelligent Control and Systems, Harbin Institute of Technology, Harbin 150080, P.R. China (e-mail: hjgao@hit.edu.cn).
}
}

\maketitle

\begin{abstract}
Missing attribute issues are prevalent in the graph learning, leading to biased outcomes in Graph Neural Networks (GNNs). Existing methods that rely on feature propagation are prone to cold start problem, particularly when dealing with attribute resetting and low-degree nodes, which hinder effective propagation and convergence. To address these challenges, we propose \textit{AttriReBoost} (ARB), a novel method that incorporates propagation-based method to mitigate cold start problems in attribute-missing graphs. ARB enhances global feature propagation by redefining initial boundary conditions and strategically integrating virtual edges, thereby improving node connectivity and ensuring more stable and efficient convergence. This method facilitates gradient-free attribute reconstruction with lower computational overhead. 
The proposed method is theoretically grounded, with its convergence rigorously established. Extensive experiments on several real-world benchmark datasets demonstrate the effectiveness of ARB, achieving an average accuracy improvement of $5.11\%$ over state-of-the-art methods. Additionally, ARB exhibits remarkable computational efficiency, processing a large-scale graph with $2.49$ million nodes in just $16$ seconds on a single GPU. Our code is available at \url{https://github.com/limengran98/ARB}.
\end{abstract}

\begin{IEEEkeywords}
Attribute-Missing Graphs, Cold Start Problem, Feature Propagation, Graph Learning.
\end{IEEEkeywords}

\section{Introduction}
The significance of graph data in representing complex networks is well-established \cite{ji2022smoothness, KUANG2024100146, li2024redundancy, yuan2024Semi, liu2024multi}, yet most real-world scenarios often suffer from missing attribute features that represent node semantic information in graphs \cite{peng2020graph, adhikari2022comprehensive, gao2023Representation, li2024scae}. Therefore, the task of reconstructing missing attributes in graphs becomes essential for comprehensive network analysis \cite{you2020handling, guo2023fair}. Recent advancements based Graph Neural Networks (GNNs) \cite{chen2022learning, yoo2022accurate, tu2022initializing} provide opportunities for effective missing attribute reconstruction. However, GNNs typically struggle with oversmoothing and high computational costs \cite{chen2020measuring}. While Feature propagation methods \cite{rossi2022unreasonable, um2023confidence} could address these issues, they still face a common challenge: \textbf{\uline{the cold start problem in reconstructing missing attributes.}}

\begin{figure}[t]
\centerline{\includegraphics[width=7.5cm]{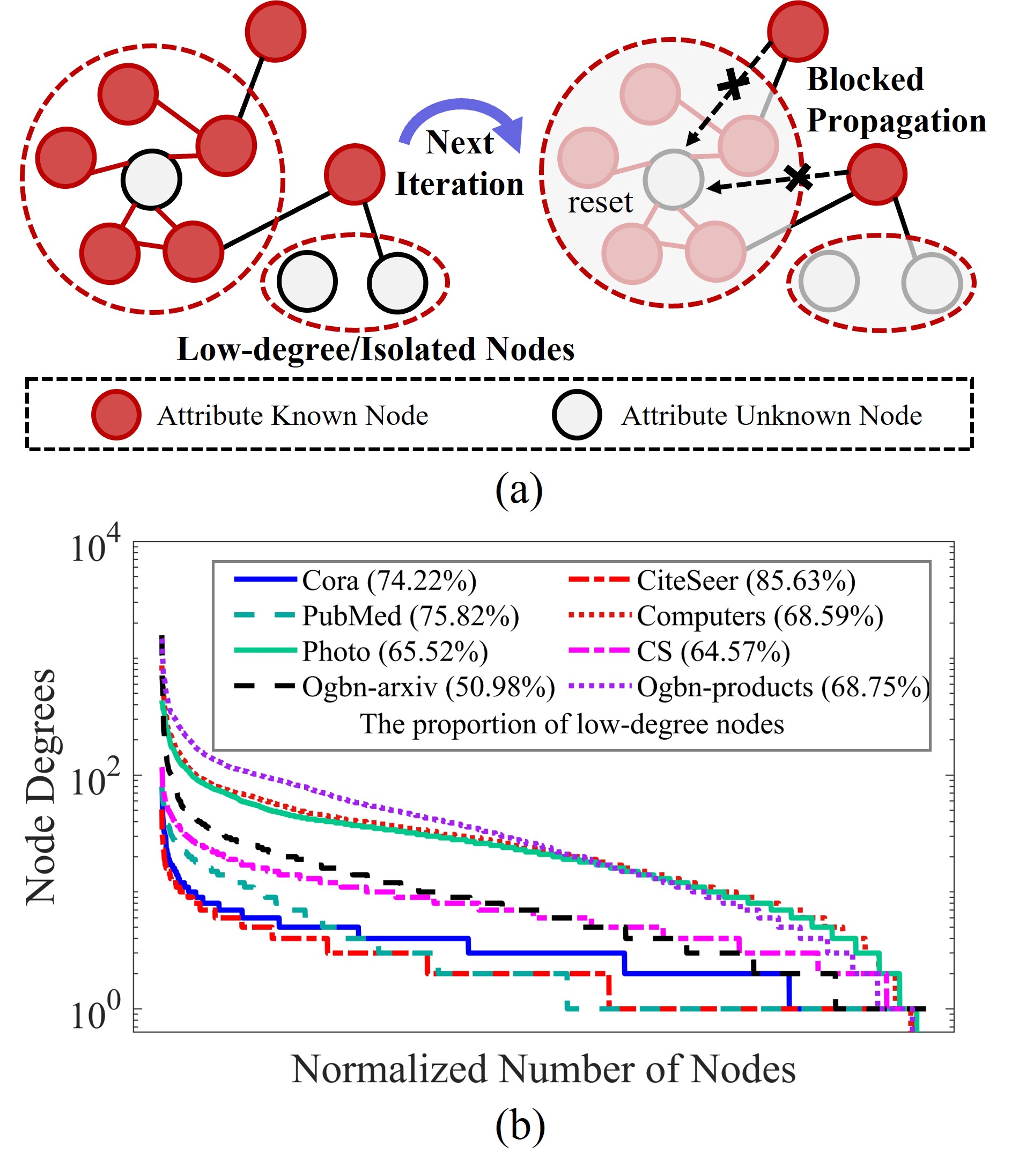}}
\caption{(a) Cold start examples in attribute-missing graphs. (b) Real-world datasets present a long-tail distribution, and tail/cold start nodes are difficult to participate in propagation.}
\label{fig1}
\end{figure}

Two key factors contribute to the cold start problem in attribute-missing graphs. First, methods like \cite{rossi2022unreasonable, um2023confidence} require resetting known nodes after each feature propagation iteration to preserve initial attributes. However, as shown in Figure \ref{fig1}(a), this resetting causes unknown nodes to repeatedly receive the same initial attributes from known nodes, hindering global information propagation. Second, while high-quality and rich connectivity can facilitate feature propagation, real-world graphs often contain many low-degree or isolated nodes, leading to a long-tail distribution of node degrees \cite{zheng2021cold, zhou2023Multiview}. As illustrated in Figure \ref{fig1}(b), approximately 70\% of nodes in each dataset are low-degree, making it difficult for propagation methods to effectively generalize to these tail or cold start nodes, thereby impacting the reconstruction of missing attributes.

Inspired by techniques like Feature Propagation (FP) \cite{rossi2022unreasonable} and Personalized PageRank \cite{gasteiger2018predict}, we propose a novel method named \textit{AttriReBoost} (ARB) to enhance attribute reconstruction and mitigate the cold start problem. ARB improves propagation connectivity by redefining initial boundary conditions and introducing effective virtual edges. By redefining boundary conditions, ARB dynamically transmits attributes from both nearby and distant known nodes to unknown nodes, avoiding information loss from repeated resets and ensuring continuous, updated attribute information. The virtual edges create a virtually fully connected overlay, allowing low-degree or isolated nodes to better participate in the propagation process, ensuring robust connectivity for convergence.

Building on these ideas, we establish an optimization function and derive ARB's iterative algorithm, rigorously proving its convergence using the \textit{Banach-Fixed Point Theorem}. ARB's gradient-free propagation eliminates the computational load of backpropagation and gradient learning, introducing only two additional hyperparameters compared to FP \cite{rossi2022unreasonable}, thus achieving attribute reconstruction with lower computational cost. Experimental validation shows that ARB outperforms state-of-the-art methods in both feature propagation and node classification. Compared to FP and Stochastic Gradient Descent (SGD) methods \cite{herbert1951stocha}, ARB achieves faster convergence and early stopping, making it a more efficient choice for large-scale and cold start-sensitive applications.

The innovations of this paper are summarized as follows:
\begin{itemize}
\item To address the cold start problem in attribute-missing graphs, ARB introduces innovative techniques that redefine boundary conditions and integrate virtual edges into the feature propagation process. Grounded in both empirical insights and rigorous mathematical theory, these advancements enhance node connectivity and significantly improve propagation efficiency.

\item ARB enhances the efficiency of graph-based learning by utilizing a gradient-free message passing framework, which significantly accelerates convergence and enables early stopping. This method simplifies the computational process while enabling more efficient and scalable processing of large-scale graphs.

\item Extensive experiments on real-world datasets demonstrate the superiority of ARB over state-of-the-art methods, \uline{showing an average accuracy improvement of 5.11\%}. Additionally, ARB demonstrates significant computational efficiency, \uline{processing large-scale graphs with 2.49 million nodes in just 16 seconds on a single GPU}, achieving faster and more stable convergence. which is of significant value in real-world applications where computational resources and time are critical.
\end{itemize}

This paper is structured as follows: Section \ref{s2} provides a review of related work, Section \ref{s3} outlines the problem, Section \ref{s4} introduces the proposed methods, Section \ref{s5} presents the experimental results, and Section \ref{s6} concludes with a discussion of potential future research.

\section{Related Work}\label{s2}

\subsection{Attribute Missing Graph Learning}
In the early work, \cite{csimcsek2008navigating} utilized mean pooling to aggregate the features of neighboring nodes. \cite{cai2010singular} proposed the Singular Value Thresholding (SVT) algorithm, which completed matrix imputation by adjusting singular values. Besides, incomplete multi-view learning \cite{wen2020incomplete, wen2021Generalized, xia2022Tensor, Yu2024Differentiated} has been widely studied as it addresses the challenge of missing or incomplete data across multiple views, enabling the development of robust models that can leverage information from different perspectives even when some data is missing or corrupted.

\textbf{GNN based Methods}  With the advent of deep learning, \cite{yoon2018gain, you2020handling} used GNNs to generate missing data. \cite{huang2019graph} and \cite{chen2019attributed} employed attributed random walk techniques to create nodes embedded on bipartite graphs with node attributes. \cite{spinelli2020missing} proposed graph denoising autoencoders, where each edge encoded the similarity between patterns to complete missing attributes. \cite{taguchi2021graph} transformed missing attributes into Gaussian mixture distribution, enabling Graph Convolutional Networks (GCN) \cite{kipf2016semi} to be applied to incomplete network attributes. SAT \cite{chen2022learning} and SVGA \cite{yoo2022accurate} employed separate subnetworks for nodes attributes and graph structure to impute missing data with structural information, using GANs and Graph Markov Random Fields (GMRFs) \cite{lauritzen1988local} respectively, guided by shared latent space assumptions. Amer \cite{jin2022amer} introduced a unified framework that combines attribute completion and embedding learning, leveraging mutual information maximization and a novel GAN-based attribute-structure relationship constraint to improve performance. ITR  proposed by \cite{tu2022initializing} initially filled in missing attributes using the graph's structural information, then adaptively refined the estimated latent variables by combining observed attributes and structural information. MAGAE \cite{gao2023handling}, on the other hand, employed a regularized graph autoencoder that mitigates spectral concentration issues by maximizing graph spectral entropy, enhancing the imputation of missing attributes. For community detection in attribute-missing networks, CAST \cite{li2024csat} adopted a Transformer-based architecture, integrating contrastive learning, sampling, and propagation strategies to effectively capture node relationships and address missing attribute challenges.\cite{peng2024multi} suggested imputing attributes in the input space by leveraging parameter initialization and graph diffusion to generate multi-view information.

\textbf{Propagation based Methods} In addition to the aforementioned deep learning-based methods, some propagation-based methods have been widely focused on due to their low complexity and high scalability. \cite{rossi2022unreasonable} introduced the FP method, which reconstructed missing features by minimizing Dirichlet energy and diffusing known features across the graph structure. \cite{um2023confidence} proposed the Pseudo-Confidence Feature Imputation (PCFI) method, which enhanced feature propagation by incorporating a pseudo-confidence-based weighting mechanism during propagation. 

\subsection{Cold Start Problem} 

The cold start problem, which arises in scenarios such as recommendation systems and information pushing, is a significant challenge due to the lack of sufficient user or item data at the initial stages. To mitigate this issue, several approaches have been proposed. For instance, \cite{hao2021pre} and \cite{gong2023unified} explored transfer learning methods to leverage knowledge from related tasks or domains, aiming to improve the performance of models with limited data. Additionally, \cite{cai2023user} combined multi-task learning for graph pre-training, allowing for the transfer of useful representations across different tasks to enhance the model's ability to generalize from sparse information. Meanwhile, \cite{yang2021extract} and \cite{huang2023aligning} focused on specialized distillation methods to transfer knowledge from well-trained models to improve the training of models with insufficient data. However, despite the success of these approaches in addressing cold start problems, they do not specifically tackle the issue of attribute-missing graphs, where the challenge lies in recovering missing or incomplete node attributes while maintaining graph structure and learning performance.

\subsection{Summary} 
Although GNN-based methods generally perform well with attribute-missing graphs, they suffer from high computational complexity and resource demands, limiting their scalability. Additionally, GNNs are prone to the oversmoothing problem, where node representations become indistinguishable as layers increase. In contrast, propagation-based methods are simpler and more scalable but struggle with the cold start problem, as they rely heavily on existing graph structure and known attributes, leading to suboptimal feature reconstruction and difficulty in capturing nuanced relationships.

\section{Preliminaries}\label{s3}
\subsection{Problem Definition}
We define an attribute-missing graph $G=(\mathcal{V}, \mathcal{E}, \mathbf X, k)$, where $\mathcal{V} = \mathcal{V}_k \cup \mathcal{V}_u$. $\mathcal{V}_k$ and $\mathcal{V}_u$ denote the sets of nodes with known and unknown (missing) attribute features, respectively. The node attribute matrix is denoted by $\mathbf X \in \mathbb{R}^{N\times F}$. For the total nodes $N$, only nodes $k$ possess attributes. The adjacency matrix is $\mathbf A \in \{0,1\}^{N\times N}$, diagonal degree matrix is $\mathbf D$, symmetric normalized adjacency matrix is $\widetilde {\mathbf A} = \mathbf D^{-1/2}\mathbf{AD}^{-1/2}$, $\widetilde {\mathbf a}_{ij}$ represent the individual elements of $\widetilde {\mathbf A}$, and symmetric normalized Laplacian matrix $\mathbf L=\mathbf I-\widetilde {\mathbf A}$. The goal of this work is to reconstruct the missing attribute feature and apply it to downstream classification tasks.

\subsection{Feature Propagation}
To reconstruct the missing attributes $\mathbf X_u$ based on the known attribute features $\mathbf X_k$ and the graph $G$, the optimization function of Feature Propagation (FP) \cite{rossi2022unreasonable} can be expressed as a process of minimizing the Dirichlet energy \cite{aubert1997variational, loncaric1998survey}. For distinction, define the matrix $\mathbf Z$ as the initial attribute feature matrix of the graph, encompassing both known and unknown node attributes. The optimization function can be expressed as:
\begin{equation}\label{e1}
\min _{\mathbf{X}} \mathcal{L}\!=\!\sum_{({i}, {j}) \in \mathcal{E}} \tilde{\mathbf{a}}_{{ij}}\left(\mathbf{x}_{{i}}-\mathbf{x}_{{j}}\right)^2=\operatorname{tr}\left(\mathbf{X}^{\top} \mathbf{L} \mathbf{X}\right). \text { s.t. } \mathbf{X}_{{k}}=\mathbf{Z}_{{k}}
\end{equation}
Solving the optimization function yields:
\begin{equation}\label{e2}
\nabla \mathcal{L}(\mathbf{X})=\mathbf{L X}=(\mathbf{I}-\widetilde{\mathbf{A}}) \mathbf{X}=0. \quad \text { s.t. } \mathbf{X}_{{k}}=\mathbf{Z}_{{k}}
\end{equation}
Initially, we set ${\mathbf{X}}^{(0)} = \mathbf{Z}$. The number of iterations is defined as $l$, The attributes are updated by $\widetilde {\mathbf{A}}$, and the known nodes are reinitialize after current propagation:
\begin{equation}\label{e3}
\left\{
\begin{aligned}
& {\mathbf X}^{(l+1)} = \widetilde {\mathbf A}{\mathbf X}^{(l)} && \triangleright \text{ Propagation}\\
& {\mathbf X}_k^{(l+1)} = {\mathbf Z}_k \phantom{= \widetilde {\mathbf A}{\mathbf X}^{(l)}}&& \triangleright \text{ Reset}
\end{aligned}
\right.
\end{equation}
Although traditional feature propagation methods lay a foundation for reconstructing missing features, they struggle to address the cold start problem. Motivated by this, the following section introduces the proposed ARB.

\section{Proposed Method}\label{s4}
\begin{figure}[h]
\centering
\includegraphics[width=3.5in]{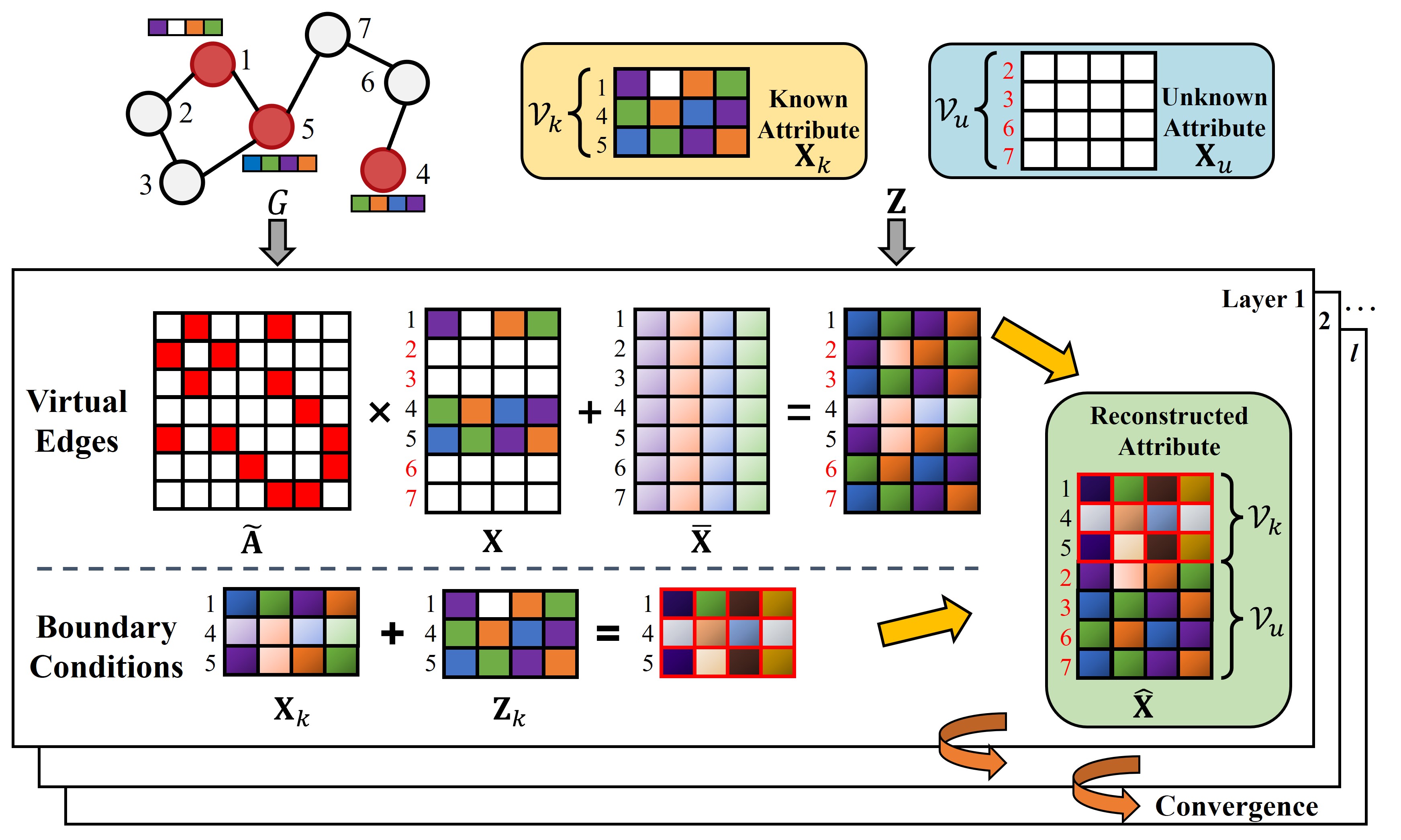}
\caption{The overall framework of ARB. ARB address the cold start problem in attribute-missing graph learning by enhancing node connectivity through virtual edges and redefining initial boundary conditions. ARB boosts propagation-based methods to accurately and effectively recover missing attributes.}
\label{fig2}
\end{figure}

A crucial prerequisite for attribute-missing graph learning is the accurate and effective recovery of missing attributes. However, the cold start problem worsen recovery, as low-degree and isolated nodes in the graph can hinder iteration convergence. This paper proposes a novel \textit{AttriReBoost} (ARB) method. To address the cold start problem, ARB boosts propagation based methods, redefines the initial boundary conditions of FP, and establishes virtual edges to enhance node connectivity, reconstructs missing attributes accurately and effectively. The overall framework is illustrated in Figure \ref{fig2}.

\subsection{Redefinition of Boundary Conditions}
An intuitive method to reconstructing missing attributes is to converge to an optimal value by iterative approximating during propagation. However, the repetitive resetting of known node attributes in propagation disrupts the smooth flow of information. As shown in Figure \ref{tab_1}(a), for unknown nodes, they receive same information stems from the initial attributes $\mathbf{Z}_k$ of known nodes during each iteration, if the majority of their neighbors are known nodes, hindering the propagation of global information. To address this, ARB redefines the boundary conditions of the optimization Equation \eqref{e1} and dynamically adjust the initialization of known nodes. 
\paragraph*{Statement 1} \textit{\uline{The New Boundary Conditions term in Equation \eqref{e4} incorporates known node attributes into the optimization process. This term is expressed in Lagrangian form as a sum of squared differences, weighted by the parameter $\eta$.}}
\begin{equation}\label{e4}
\begin{aligned}
\min_\mathbf X \mathcal{L} &= \underbrace{\sum_{(u,v)\in \mathcal E} \widetilde {\mathbf a}_{uv}(\mathbf x_u-\mathbf x_v)^2}_{\text{Feature Propagation}} + \underbrace{\eta\sum_{v \in \mathcal{V}_k}(\mathbf x_v-\mathbf z_v)^2}_{\text{New Boundary Conditions}}\\
&= \text{tr}(\mathbf X^\top\mathbf{LX}+\eta(\mathbf X_k-\mathbf Z_k)^\top(\mathbf X_k-\mathbf Z_k)).
\end{aligned}
\end{equation}
\subsection{Virtual Edges for Connectivity}
In real-world graphs, many nodes are sparsely connected, and some are even isolated, resulting in a long-tail distribution of node degrees. Consequently, it struggles to effectively generalize to tail or cold start nodes for feature propagation. Observing the iterative Equation \eqref{e3}, $\mathbf{X}_k^{\left(l+1\right)}=\widetilde{\mathbf{A}}\mathbf{X}^{\left(l\right)}$, since $\widetilde{\mathbf{A}}$ is a symmetric normalized matrix, its spectral radius is not strictly less than 1, which could not guarantee convergence. To address this issue, a fully connected graph $\overline{G}=\left(\mathcal{V},\overline{\mathcal{E}},\mathbf{Z},k\right)$ is introduced, where the virtual edge set $\overline{\mathcal{E}}$ assumes that all node pairs are connected. 

\paragraph*{Statement 2}\textit{\uline{Extending the New Boundary Conditions term, the Virtual Edges term in Equation \eqref{e5} introduces additional connectivity via $\overline{\mathcal{E}}$ to enhance global information. This term is represented as Dirichlet energy, incorporated as a penalty term, weighted by the parameter $\theta$.}}

\begin{equation}\label{e5}
\begin{aligned}
&\min_\mathbf X \mathcal{L} = \underbrace{\sum_{(u,v)\in \mathcal{E}} \widetilde {\mathbf a}_{uv}(\mathbf x_u-\mathbf x_v)^2}_{\text{Feature Propagation}}  + \underbrace{\eta\sum_{v\in \mathcal{V}_k}(\mathbf x_v-\mathbf z_v)^2}_{\text{New Boundary Conditions}} \\
&+ \underbrace{\theta\sum_{(i,j)\in \mathcal{\overline E}} \overline {\mathbf a}_{ij}(\mathbf x_i - \mathbf x_j)^2}_{\text{Virtual Edges}} \\
&= \text{tr}\left(\mathbf{X}^\top\mathbf{LX}   + \eta(\mathbf X_k-\mathbf Z_k)^\top(\mathbf X_k-\mathbf Z_k) + \theta \mathbf X^\top\mathbf L_1\mathbf X \right),
\end{aligned}
\end{equation}
where $\overline {\mathbf a}_{ij}$ is the element of the normalized adjacency matrix of $\overline{G}$, $\mathbf{L}_1=\frac{N}{N-1}\mathbf{I}-\frac{1}{N-1}\mathbf{J}$, and $\mathbf{J}$ is an all-ones matrix.
\subsection{Optimization Function of ARB}
To optimize the Equation \eqref{e5}, we first define the matrix $\mathbf I_k^0 = \text{diag}(\{\lambda_1,\lambda_2,\cdots,\lambda_N\}),
\lambda_i = \begin{cases}
1 &\text{ if } i\in\mathcal{V}_k \\
0 &\text{otherwise}
\end{cases}$, representing the known and unknown nodes, and explores the gradient $\nabla\mathcal{L}(\mathbf X)$:

\begin{equation}\label{e6}
\resizebox{1\linewidth}{!}{$
\begin{aligned}
& \nabla\mathcal{L}(\mathbf X) = \mathbf {LX} + \eta(\mathbf X_k-\mathbf Z_k) + \theta \mathbf L_1 \mathbf X = 0 \\
\implies& (\mathbf I+\eta \mathbf I_k^0 + \theta \mathbf I)\mathbf X = \widetilde {\mathbf A}\mathbf X+\eta \mathbf I_k^0\mathbf Z + \theta (\frac{N}{N-1}\overline {\mathbf X} - \frac{1}{N-1}\mathbf X) \\
\implies& (\frac{\theta N+N-1}{N-1}\mathbf I+\eta \mathbf I_k^0)\mathbf X = \widetilde {\mathbf A}\mathbf X+\eta \mathbf I_k^0\mathbf Z + \frac{\theta N}{N-1}\overline {\mathbf X},
\end{aligned}$}
\end{equation}
where $\overline {\mathbf X}$ represents the mean of $\mathbf X$. Considering unknown nodes and known nodes separately, Equation \eqref{e6} is further simplified to: 
\begin{equation}\label{e7}
\resizebox{1.05\linewidth}{!}{$
\begin{aligned}
& \begin{bmatrix} \mathbf X_k \\ \mathbf X_u \end{bmatrix} = \begin{bmatrix} \left[\frac{\eta}{\frac{\theta N+N-1}{N-1}+\eta}\mathbf Z + \frac{1}{\frac{\theta N+N-1}{N-1}+\eta}\widetilde {\mathbf A}\mathbf X + \frac{\frac{\theta N}{N-1}}{\frac{\theta N+N-1}{N-1}+\eta}\overline {\mathbf X} \right]_k \\ [\frac{N-1}{\theta N+N-1}\widetilde {\mathbf A}\mathbf X+\frac{\theta N}{\theta N+N-1}\overline {\mathbf X}]_u \end{bmatrix}.
\end{aligned}$}
\end{equation}

Let $\alpha = \frac{N-1}{\theta N+N-1}$ and $\beta = \frac{1/\alpha}{1/\alpha+\eta}$, both of which belong to the open interval $(0,1)$, we have:
\begin{equation}\label{e8}
\begin{aligned}
\begin{bmatrix} \mathbf X_k \\ \mathbf X_u \end{bmatrix} = \begin{bmatrix} [((1-\beta)\mathbf Z + \beta(\alpha\widetilde {\mathbf A}\mathbf X + (1-\alpha)\overline {\mathbf X})]_k \\ [\alpha\widetilde {\mathbf A}\mathbf X+(1-\alpha)\overline {\mathbf X}]_u \end{bmatrix}.
\end{aligned}
\end{equation}

We then relate ARB recursively as follows:
\begin{equation}\label{e9}
\left\{
\begin{aligned}
& {\mathbf X}^{(l+1)}  = \alpha\widetilde {\mathbf A}\mathbf X^{(l)}  + (1-\alpha) \overline {\mathbf X}^{(l)}  &&\triangleright \text{ Global Propagation}\\
& {\mathbf X}_k^{(l+1)}  = \beta \mathbf X_k^{(l)}  + (1-\beta)\mathbf Z_k &&\triangleright \text{  Moving Reset}
\end{aligned}
\right.
\end{equation}
ARB process is shown in Algorithm \ref{aa1}.
\begin{algorithm}[h]
\caption{AttriReBoost}
\label{aa1}
\begin{algorithmic}[1]  
\REQUIRE known attribute matrix $\mathbf Z_k$, normalized adjacency matrix $\widetilde {\mathbf A}$, hyperparameters $\alpha$, $\beta$ and $l$
\STATE  $\mathbf X \leftarrow 0, \mathbf X_k \leftarrow \mathbf Z_k$
\FOR{$l$ iterations}
    \STATE ${\mathbf X} \leftarrow \alpha \widetilde {\mathbf A}\mathbf X + (1-\alpha)\overline {\mathbf X} \qquad\triangleright\text{Global Propagation}$
    \STATE ${\mathbf X}_k \leftarrow \beta \mathbf X_k +(1-\beta)\mathbf Z_k \qquad\triangleright\text{ Moving Reset}$
\ENDFOR
\STATE \textbf{Until} $\mathbf X$ convergence
\ENSURE Reconstructed attributes $\hat{\mathbf X}_{u}$
\end{algorithmic}
\end{algorithm}

\subsection{ARB Convergence and Steady State Proof}\label{Sec:IV4}
The ultimate goal is to demonstrate that the proposed ARB can iteratively approach the optimal value to achieve attribute reconstruction. By introducing metric space, the convergence of the proposed ARB has been rigorously proven at the theoretical level.

$\square$ \textit{Convergence Proof:} Since $\widetilde {\mathbf A}$ be the symmetric normalized adjacency matrix, $\rho(\widetilde {\mathbf A}) \leq 1$ \cite{chung1997spectral}. Let $\mathbf B = \alpha \widetilde {\mathbf A} + (1-\alpha)\frac{1}{N}\mathbf J$ is a strong connection matrix (irreducide matrix), $\rho(\mathbf B) = \Vert \mathbf B \Vert_2 \le \alpha \Vert \widetilde{\mathbf A} \Vert_2 + (1-\alpha)\Vert \mathbf J \Vert_2 = \alpha\rho(\widetilde{\mathbf A})+(1-\alpha)\rho(\mathbf J) \le 1$.

Equation \eqref{e9} can be written as:
\begin{equation}\label{pe1}
\begin{bmatrix} \mathbf X_k \\ \mathbf X_u \end{bmatrix} = \begin{bmatrix} \beta \mathbf B_{kk} & \beta \mathbf B_{ku} \\ \mathbf B_{uk} & \mathbf B_{uu} \end{bmatrix}\begin{bmatrix} \mathbf X_k \\ \mathbf X_u \end{bmatrix} + \begin{bmatrix} (1-\beta)\mathbf Z_k \\ 0 \end{bmatrix}.
\end{equation}

Let $\mathbf K = \begin{bmatrix} \beta \mathbf B_{kk} & \beta \mathbf B_{ku} \\ \mathbf B_{uk} & \mathbf B_{uu} \end{bmatrix}, \mathbf C = \begin{bmatrix} (1-\beta)\mathbf Z_k \\ 0 \end{bmatrix}$, $0 \le \mathbf K \le \mathbf B$ elementwise. Because $\beta<1$, which means $\mathbf K \ne \mathbf B$, therefore $0 \le \mathbf K < \mathbf B$.

Given that $\mathbf B$ is a strongly connected matrix, and $\mathbf K \ge 0, \mathbf B \ge 0$, $\mathbf K+\mathbf B$ is also a strongly connected matrix and therefore irreducible.
And with $0 \le \mathbf K \le \mathbf B$ elementwise and $\mathbf K \ne \mathbf B$, we can deduce that $\rho(\mathbf K)<\rho(\mathbf B)$.
Therefore $\rho(\mathbf K)<\rho(\mathbf B)\le 1$, that is, $\rho(\mathbf K)<1$.

Additionally, if $\mathbf B$ can be expressed as a diagonal matrix composed of a series of strongly connected matrices $\mathbf B_i$. Responsive, $\mathbf K$ can also be decomposed into a series of $0 \le \mathbf K_i \le \mathbf B_i$ elementwise, and there exists at least one $\mathbf K_j$ that satisfies $\mathbf K_j \ne \mathbf B_j$, so $\rho(\mathbf K_j)<1$, and

\begin{equation}\label{e10}
\resizebox{1\linewidth}{!}{$
 \rho(\mathbf K) = \max \frac{\mathbf{X}^\top \mathbf K\mathbf{X}}{\mathbf{X}^\top  \mathbf X} = \max \frac{\sum\mathbf{x}_i^\top \mathbf K_i\mathbf x_i}{\sum\mathbf x_i^\top  \mathbf x_i} < \max \frac{\sum\mathbf x_i^\top \mathbf x_i}{\sum\mathbf x_i^\top  \mathbf x_i} = 1$}.
 \end{equation}

According Equation \eqref{pe1}, Let $f(\mathbf X) = \mathbf{KX} + \mathbf C$. Take the spectral norm $\Vert \mathbf X \Vert_2 = \sqrt{\rho(\mathbf X^\top\mathbf X)}$ as the measure.
From $\rho(\mathbf K) < 1$, we know $\rho(\mathbf K^\top \mathbf K) < 1$. And use the feature vector $\{\mathbf v_1,\mathbf v_2,...,\mathbf v_n\}$ of $\mathbf K^\top \mathbf K$ to represent $\Delta \mathbf X=\sum_i w_i \mathbf v_i$, where \(w_i\) are the coefficients. Then:

\begin{equation}\label{e11}
\begin{aligned}
&\Vert \Delta f(\mathbf X) \Vert_2 = \Vert \mathbf K\Delta \mathbf X \Vert_2 
= \Delta \mathbf X^\top  \mathbf K^\top  \mathbf K \Delta \mathbf X \\
&= \sum_{ij} w_iw_j\mathbf v_i\lambda_j\mathbf v_j 
\le \rho(\mathbf K^\top \mathbf K)\sum_{ij} w_iw_j\mathbf v_i\mathbf v_j \\
&= \rho(\mathbf K^\top \mathbf K)\Delta \mathbf X^\top \Delta \mathbf X 
=\rho(\mathbf K^\top \mathbf K)\Vert \Delta \mathbf X \Vert_2 
< \Vert \Delta \mathbf X \Vert_2.
\end{aligned}
\end{equation}

That is, $f$ is a contraction mapping, and according to \textit{Banach Fixed Point Theorem} \cite{banach1922operations}, $\mathbf X = f(\mathbf X)$ has a unique fixed-point. Therefore, the recursive expression $\mathbf X^{(l+1)} = \mathbf {KX}^{(l)}+\mathbf C$ converges to a unique value. $\blacksquare$

$\square$ \textit{Steady State Proof:}
According to Equations \eqref{e5} and \eqref{e6}, the steady state of ARB can be given as follows:
\begin{equation}
\begin{aligned}
\nabla\mathcal{L}(\mathbf X) &= \mathbf{LX} + \eta(\mathbf X_k-\mathbf Z_k) + \theta \mathbf L_1 \mathbf X = 0 \\
&\implies (\mathbf L+\eta \mathbf I_k^0 + \theta \mathbf L_1)X = \eta \mathbf Z_k \\
&\implies \mathbf X = (\mathbf L+\eta \mathbf I_k^0 + \theta \mathbf L_1)^{-1}\eta \mathbf Z_k
\end{aligned}
\end{equation}

Consider the Rayleigh quotient:
\begin{equation}
\resizebox{1.\linewidth}{!}{$
\begin{aligned}
\mathbf R &= \frac{\mathbf{X}^\top (\mathbf L+\eta \mathbf I_k^0 + \theta \mathbf L_1)\mathbf{X}}{\mathbf{X}^\top \mathbf{X}} \\
&= \frac{\sum_{(i,j)\in \mathcal{E}}\widetilde {\mathbf a}_{ij}{(\mathbf x_i-\mathbf x_j)^2} + \eta\sum_{i \in \mathcal{V}_k} \mathbf x_i^2+ \sum_{(i,j)\in \mathcal{\overline E}} \overline {\mathbf a}_{ij}(\mathbf x_i - \mathbf x_j)^2}{\sum_{i \in \mathcal{V}} \mathbf x_i^2} > 0.
\end{aligned}$}
\end{equation}

In fact, \( \mathbf{X}^\top (\mathbf{L} + \eta \mathbf{I}_k^0 + \theta \mathbf{L}_1)\mathbf{X} = 0 \iff \forall i,j \in \mathcal{V}, \, \text{s.t. } \mathbf{X}_i = \mathbf{X}_j = 0 \), but \( \mathbf{X} \ne 0 \). Thus, all eigenvalues of \( \mathbf{L} + \eta \mathbf{I}_k^0 + \theta \mathbf{L}_1 \) are positive, making the matrix invertible. $\blacksquare$

\subsection{Complexity, Scalability, and Learning}
The goal of ARB is to reconstruct missing attributes. Compared to FP, it introduces only two additional adjustable hyperparameters, $\alpha$ and $\beta$, thereby retaining all the advantages of FP in terms of low complexity $O(\vert \mathcal{E} \vert + F\vert \mathcal{V} \vert)$ and scalability \cite{rossi2022unreasonable}. ARB is a gradient-free method that can be run as preprocessing on the CPU for large graphs and integrated with any graph learning model, like GCN \cite{kipf2016semi} and GAT \cite{velivckovic2017graph}, to generate predictions for downstream tasks \cite{wu2019simplifying}.

\section{Experimentation and Analysis}\label{s5}
To thoroughly verify ARB's effectiveness in attribute reconstruction and downstream tasks, we address the following key questions:

\textbf{Q1}: How does ARB perform in attribute reconstruction?

\textbf{Q2}: How does ARB perform in downstream tasks after attribute-missing reconstruction?

\textbf{Q3}: How does ARB address the cold start problem and convergence issues?

\textbf{Q4}: Are all components of ARB effective?

\textbf{Q5}: How does ARB perform under different missing rates?

\textbf{Q6}: How is the computational efficiency of ARB?
\begin{table}[htbp]\scriptsize
  \centering
  \caption{Data statistics.}
      \renewcommand\arraystretch{1}
   \renewcommand\tabcolsep{2pt}
    \begin{tabular}{ccccccc}
    \toprule
    Dataset &{Type} & {Nodes} & {Edges} & {Feature} & {Classes} & {Subgraphs}  \\
    \hline
    Cora & Binary& 2,708 & 5,278 & 1,433 & 7 & 78 \\
    CiteSeer &  Binary&3,327 & 4,228 & 3,703 & 6 & 438\\
    Computers & Binary&13,752 & 245,861 & 767 & 10& 314 \\
    Photo &  Binary&7,650 & 119,081 & 745 & 8 & 136\\
    PubMed &  Continuous&19,717 & 44,324 & 500 & 3 & 1\\
    CS & Continuous& 18,333 & 81,894 & 6,805 & 15 & 1\\
    Ogbn-Arxiv & Continuous&169,343 & 1,166,243 & 128 & 40 & 1\\
    Ogbn-Products & Continuous&2,449,029 & 61,859,140 & 100 & 47 & 1\\
    \bottomrule
    \end{tabular}%
  \label{tab_1}%
\end{table}%

\subsection{Experimental Setup}
\subsubsection{Dataset}
We chose eight public graph datasets included Cora, CiteSeer, PubMed \cite{yang2016revisiting}, CS  \cite{shchur2018pitfalls}, Computers, Photo \cite{shchur2018pitfalls}, and large-scale datasets Ogbn-Arxiv and Ogbn-Products \cite{hu2020open} for this study, each containing vital information like nodes, edges and attribute features. Table \ref{tab_1} shows data statistics for each dataset. 

\subsubsection{Baseline}

We compare ARB with several baseline methods. NeighAgg \cite{csimcsek2008navigating} aggregates features of one-hop neighbors using mean pooling. GNN* refers to the best-performing models among GCN \cite{kipf2016semi}, GraphSAGE \cite{hamilton2017inductive}, and GAT \cite{velivckovic2017graph}. GraphRNA \cite{huang2019graph} and ARWMF \cite{chen2019attributed} are recent feature generation methods. SAT \cite{chen2022learning} uses a shared latent space for features and graph structure. Amer \cite{jin2022amer} integrates attribute completion and embedding learning using GAN-based constraints. SVGA \cite{yoo2022accurate} employs Gaussian Markov random fields for feature estimation. ITR \cite{tu2022initializing} fills missing attributes using graph structure and refines them iteratively. FP \cite{rossi2022unreasonable} uses Dirichlet energy minimization to impute features. PCFI \cite{um2023confidence} introduces pseudo-confidence for feature imputation. MAGAE \cite{gao2023handling} mitigates spectral concentration via a graph autoencoder. CAST \cite{li2024csat} is a Transformer-based method combining contrastive learning and propagation. MATE \cite{peng2024multi} enhances attribute imputation through graph diffusion and multi-view information.

\begin{figure}[htpb]
\centering
\includegraphics[width=3in]{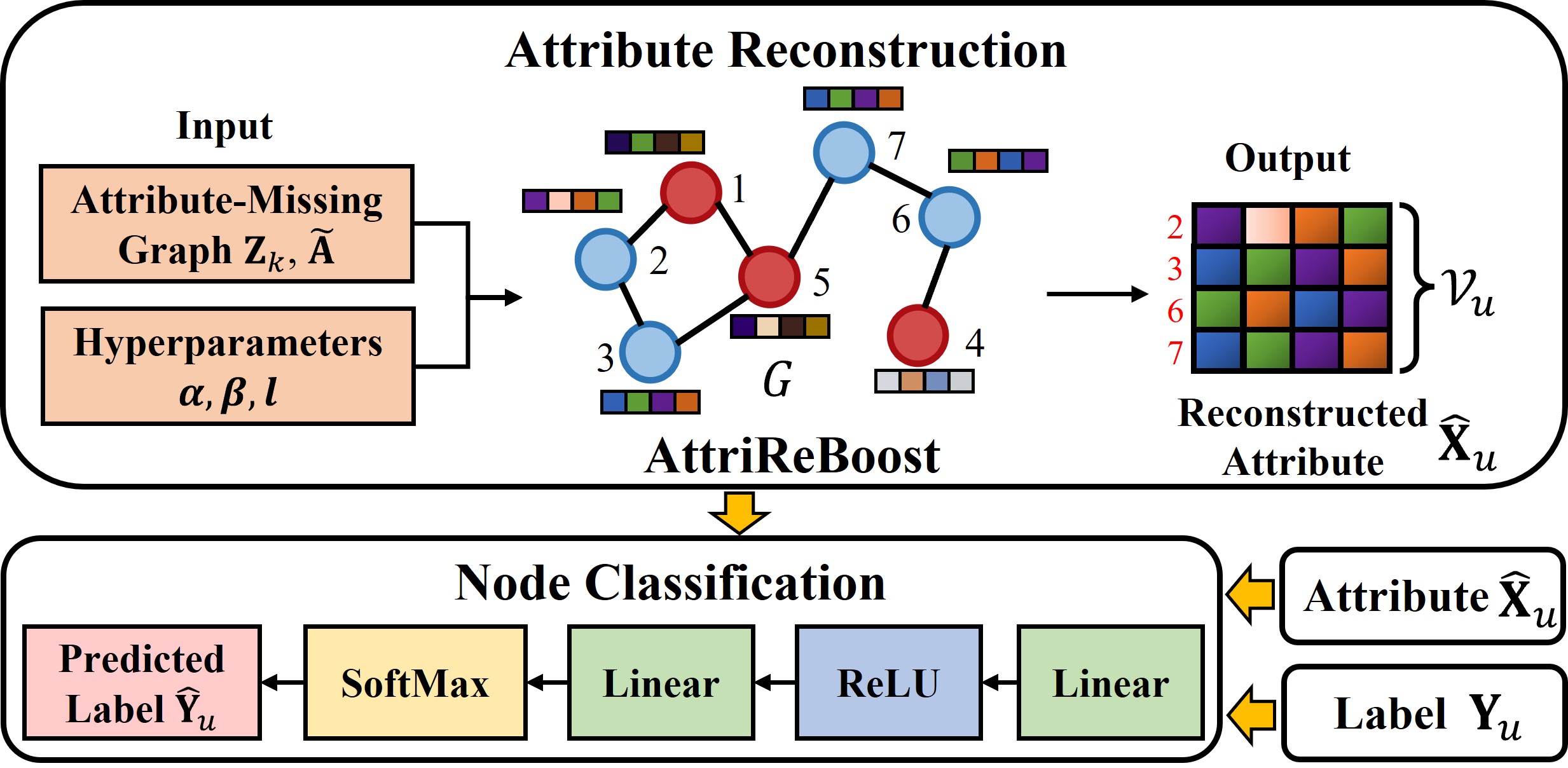}
\caption{Specific implementation process of attribute reconstruction and node classification.}
\label{figaa}
\end{figure}

\begin{table*}[htbp]
\centering
\small
  \caption{Evaluation of ARB and baseline methods on binary features for attribute reconstruction. Best results are indicated in \textcolor{blue}{\textbf{blue}}, second best results are \textcolor{darkgreen}{\textbf{green}}.}
    \renewcommand\arraystretch{1}
   \renewcommand\tabcolsep{3.8pt}
    \begin{tabular}{ccccccccccccccc}
    \toprule
    \multirow{1}[2]{*}{Metric} & \multirow{1}[2]{*}{Method} & \multirow{1}[2]{*}{Venue}& \multicolumn{3}{c}{Cora} & \multicolumn{3}{c}{CiteSeer} & \multicolumn{3}{c}{Computers} & \multicolumn{3}{c}{Photo} \\
\cline{4-15}  &   & &  @10  &  @20  &  @50  &  @10  &  @20  &  @50  &  @10  &  @20  &  @50  &  @10  &  @20  &  @50  \\
    \hline
    \multirow{10}[2]{*}{Recall} & NeighAgg & / &0.0906  & 0.1413  & 0.1961  & 0.0511  & 0.0908  & 0.1501  & 0.0321  & 0.0593  & 0.1306  & 0.0329  & 0.0616  & 0.1361  \\
      & GNN* & /&0.1350  & 0.1812  & 0.2972  & 0.0620  & 0.1097  & 0.2058  & 0.0273  & 0.0533  & 0.1278  & 0.0295  & 0.0573  & 0.1324  \\
      & GraphRNA & KDD'19& 0.1395  & 0.2043  & 0.3142  & 0.0777  & 0.1272  & 0.2271  & 0.0386  & 0.0690  & 0.1465  & 0.0390  & 0.0703  & 0.1508  \\
      & ARWMF& NeurIPS'19  & 0.1291  & 0.1813  & 0.2960  & 0.0552  & 0.1015  & 0.1952  & 0.0280  & 0.0544  & 0.1289  & 0.0294  & 0.0568  & 0.1327  \\
      & SAT & TPAMI'22 & 0.1653  & 0.2345  & 0.3612 & 0.0811  & 0.1349  & 0.2431  & 0.0421  & 0.0746  & 0.1577  & 0.0427  &0.0765  & 0.1635  \\
    & Amer& TCYB'22  & 0.1584	&0.2220&	0.3368	&0.0814	&0.1333&	0.2248&	0.0425	&0.0739	&0.1523&	0.0435&	0.0772&	0.1617 \\ 
    & SVGA &KDD'22 & 0.1718  & \textcolor{darkgreen}{\textbf{0.2486}}  &  \textcolor{darkgreen}{\textbf{0.3814}} & {0.0943}  &{0.1539}  & \textcolor{darkgreen}{\textbf{0.2782}}  & {0.0437}  & {0.0769}  & 0.1602  & \textcolor{darkgreen}{\textbf{0.0446}}  & \textcolor{darkgreen}{\textbf{0.0798}} & \textcolor{darkgreen}{\textbf{0.1670}}  \\
    & ITR & IJCAI'22& 0.1656 & 	0.2372 & 	0.3652& 0.0972&  	0.1552 & 	0.2679 	& 0.0446 & 	0.0780&  0.1530 & 	0.0434 	& 0.0778 	& 0.1635 \\
    & FP  &LOG'22& 0.1620  & 0.2268  & 0.3406  & 0.0850  & 0.1380  & 0.2311  & 0.0425  & 0.0741  & 0.1544  & 0.0434  & 0.0773  & 0.1627 \\
    & PCFI &ICLR'23 & 0.1609  & 0.2261  & 0.3434  & 0.0812  & 0.1342  & 0.2294  & 0.0429  & 0.0750  & 0.1547  & 0.0436  & 0.0774  & 0.1618 \\
    &MEGAE	&AAAI'23 &0.1730 	&0.2355 	&0.3660 &	0.0958 	&0.1497 	&0.2619 	&0.0431 &	0.0748 &	0.1543 &	0.0433 	&0.0768 	&0.1610 \\
    &CAST&TCSS'24&	 0.1720 &	0.2475 	&0.3703 &	0.0949 &	0.1506& 	0.2577 &	0.0427& 	0.0749& 	0.1558 	&0.0438 &	0.0774 	&0.1623 \\
    & MATE & IF'24& \textcolor{darkgreen}{\textbf{0.1731}} &	{0.2460}  &	0.3768  &	\textcolor{darkgreen}{\textbf{0.1000}}&  	\textcolor{darkgreen}{\textbf{0.1589}}  &	{0.2716}  &	\textcolor{darkgreen}{\textbf{0.0447}} &	\textcolor{darkgreen}{\textbf{0.0782}} 	&\textcolor{darkgreen}{\textbf{0.1618}} 	&{0.0442} &	{0.0795}& {0.1649} \\
      & ARB & Ours& \textcolor{blue}{\textbf{0.1856}} & \textcolor{blue}{\textbf{0.2599}} & \textcolor{blue}{\textbf{0.3851}} & \textcolor{blue}{\textbf{0.1046}} & \textcolor{blue}{\textbf{0.1643}} & \textcolor{blue}{\textbf{0.2823}} & \textcolor{blue}{\textbf{0.0449}} & \textcolor{blue}{\textbf{0.0784}} & \textcolor{blue}{\textbf{0.1627}} & \textcolor{blue}{\textbf{0.0455}} & \textcolor{blue}{\textbf{0.0804}} & \textcolor{blue}{\textbf{0.1681}} \\
    \hline
    \multirow{10}[2]{*}{nDCG} & NeighAgg  & /  & 0.1217  & 0.1548  & 0.1850  & 0.0823  & 0.1155  & 0.1560  & 0.0788  & 0.1156  & 0.1923  & 0.0813  & 0.1196  & 0.1998  \\
      & GNN* & /&  0.1791  & 0.2099  & 0.2711  & 0.1026  & 0.1423  & 0.2049  & 0.0673  & 0.1028  & 0.1830  & 0.0712  & 0.1083  & 0.1896  \\
      & GraphRNA& KDD'19 &  0.1934  & 0.2362  & 0.2938  & 0.1291  & 0.1703  & 0.2358  & 0.0931  & 0.1333  & 0.2155  & 0.0959  & 0.1377  & 0.2232  \\
      & ARWMF& NeurIPS'19 & 0.1824  & 0.2182  & 0.2776  & 0.0859  & 0.1245  & 0.1858  & 0.0694  & 0.1053  & 0.1851  & 0.0727  & 0.1098  & 0.1915  \\
      & SAT & TPAMI'22& 0.2250  & 0.2723  & 0.3394  & 0.1385  & 0.1834  & 0.2545  & 0.1030  & 0.1463  & 0.2346  & 0.1047  & 0.1498  & 0.2421  \\
    & Amer & TCYB'22& 0.2254 &	0.2690 &	0.3301 	&0.1364 	&0.1817 &	0.2415 	&0.1048 &	0.1468 &	0.2307 &	0.1062& 	0.1513 &	0.2408  \\
      & SVGA &KDD'22 &  \textcolor{darkgreen}{\textbf{0.2381}}  & \textcolor{darkgreen}{\textbf{0.2894}}  & \textcolor{darkgreen}{\textbf{0.3601}}  & {0.1579}  & {0.2076}  &0.2892  & {0.1068}  &{0.1509}  & {0.2397}  & 0.1084  & 0.1549  & \textcolor{darkgreen}{\textbf{0.2472}}  \\
    & ITR & IJCAI'22& 0.2288 &	0.2770 &	0.3448 & 0.1645 &	0.2129 &	0.2870 &0.1086 &	\textcolor{darkgreen}{\textbf{0.1612}} &	0.2415 	&0.1069 &	0.1526 &	0.2440 \\
    & FP&LOG'22 & 0.2306  & 0.2755  & 0.3359  & 0.1427  & 0.1891  & 0.2497  & 0.1063  & 0.1491  & 0.2345  & 0.1066  & 0.1514  & 0.2411  \\
    & PCFI &ICLR'23 & 0.2272  & 0.2723  & 0.3342  & 0.1367  & 0.1829  & 0.2445  & 0.1061  & 0.1487  & 0.2341  & 0.1063  & 0.1511  & 0.2406 \\
            &MEGAE	&AAAI'23& 0.2440 	& 0.2858 	& 0.3552 	& 0.1611 	& 0.2062 	& 0.2798 	& 0.1063 	& 0.1487 	& 0.2337 & 	0.1063 & 	0.1507 & 	0.2401 \\
        & CAST&TCSS'24	& 0.2401 	& 0.2877 & 	0.3505 & 	0.1606&  	0.2071 & 	0.2771 	& 0.1044&  	0.1447 & 	0.2335 & 	0.1062 & 	0.1510 & 	0.2410 \\
     & MATE& IF'24  & {0.2373} &	{0.2861} 	&{0.3550} &	\textcolor{darkgreen}{\textbf{0.1720}} &	\textcolor{darkgreen}{\textbf{0.2212}} &	\textcolor{darkgreen}{\textbf{0.2950}} &	\textcolor{darkgreen}{\textbf{0.1090}} 	&0.1535 	&\textcolor{darkgreen}{\textbf{0.2424}} 	&\textcolor{darkgreen}{\textbf{0.1086}} 	&\textcolor{darkgreen}{\textbf{0.1553}}& {0.2465} \\
      & ARB &Ours&  \textcolor{blue}{\textbf{0.2594}} & \textcolor{blue}{\textbf{0.3043}} & \textcolor{blue}{\textbf{0.3749}} & \textcolor{blue}{\textbf{0.1768}} & \textcolor{blue}{\textbf{0.2267}} & \textcolor{blue}{\textbf{0.3043}} & \textcolor{blue}{\textbf{0.1107}} & \textcolor{blue}{\textbf{0.1553}} & \textcolor{blue}{\textbf{0.2450}} & \textcolor{blue}{\textbf{0.1104}} & \textcolor{blue}{\textbf{0.1565}} & \textcolor{blue}{\textbf{0.2493}}\\
    \bottomrule
    \end{tabular}%
      \label{tab_2}%
\end{table*}
\begin{table}[htbp]
\centering
\small
      \caption{Evaluation of ARB and baseline methods on continuous features for attribute reconstruction. Best results are  \textcolor{blue}{\textbf{blue}}, second best results are \textcolor{darkgreen}{\textbf{green}}.}
        \renewcommand\arraystretch{1}
   \renewcommand\tabcolsep{3.5pt}
    \begin{tabular}{cccccc}
    \toprule
    \multirow{2}[1]{*}{Method} & \multirow{2}[1]{*}{Venue}& \multicolumn{2}{c}{PubMed} & \multicolumn{2}{c}{CS} \\
    \cline{3-6}  & & RMSE & CORR & RMSE & CORR \\
    \midrule
    NeighAgg &/& 0.0186  & -0.2133  & 0.0952  & -0.2279  \\
    GNN*  &/& 0.0168  & -0.0010  & 0.0850  & 0.0179  \\
    GraphRNA& KDD'19  & 0.0172  & -0.0352  & 0.0897  & -0.1052  \\
    ARWMF& NeurIPS'19 & 0.0165  & 0.0434  & 0.0827  & 0.0710  \\
    SAT& TPAMI'22   & 0.0165  & {0.0378}  &{0.0820}  & {0.0958}  \\
    Amer& TCYB'22 & 0.0185  & {0.0123}  &{0.0826}  & {0.0233}\\
    SVGA&KDD'22   & 0.0166  & 0.0280  & 0.0824  & 0.0740  \\
    ITR& IJCAI'22  & \textcolor{darkgreen}{\textbf{0.0164}}  & 0.0324  & 0.0832  & 0.0543  \\
    FP&LOG'22 & 0.0165 & \textcolor{darkgreen}{\textbf{0.0778}} &  \textcolor{darkgreen}{\textbf{0.0798}} & \textcolor{darkgreen}{\textbf{0.1657}} \\
    PCFI &ICLR'23 &  0.0166 &  0.0729  &  0.0883 & 0.1571 \\
    MAGAE&AAAI'23& 0.0185 & -0.1010 &0.0854 & 0.1656\\
    CSAT&TCSS'24& 0.0165 & 0.0710 &0.0886 & 0.1101\\
    MATE&IF'24&  0.0165 &  0.0345  &  0.0832 & 0.0645 \\
    ARB &Ours & \textcolor{blue}{\textbf{0.0161}}  & \textcolor{blue}{\textbf{0.0981}}  & \textcolor{blue}{\textbf{0.0777}}  & \textcolor{blue}{\textbf{0.1811}}  \\
    \midrule
    Method &Venue&  \multicolumn{2}{c}{Ogbn-Arxiv} & \multicolumn{2}{c}{Ogbn-Products} \\
    \midrule
    FP& LOG'22&\textcolor{darkgreen}{\textbf{0.1101}} & \textcolor{darkgreen}{\textbf{0.0028}} &  \textcolor{darkgreen}{\textbf{0.5847}} & \textcolor{darkgreen}{\textbf{0.1574}} \\
    PCFI& ICLR'23& 0.1101 & 0.0030 &  0.5848 & \textcolor{darkgreen}{\textbf{0.1574}} \\
    ARB& Ours&\textcolor{blue}{\textbf{0.1002}} & \textcolor{blue}{\textbf{0.0170}} &  \textcolor{blue}{\textbf{0.5840}} & \textcolor{blue}{\textbf{0.1577}} \\
    \bottomrule
    \end{tabular}%
  \label{tab_21}%
\end{table}

\subsubsection{Implementation Details}
Unless otherwise stated, we allocate 40\% of the observable data as the training set and consider 60\% of the attribute-missing nodes as target nodes. We split target nodes into validation and test sets in a 1:5 ratio, consistent with previous work \cite{yoo2022accurate, tu2022initializing}. ARB performs missing attribute reconstruction according to the process outlined in Algorithm \ref{aa1}, executing \(l\) iterations. In each iteration, the neighborhood information of each node is first gathered through 3rd line in Algorithm \ref{aa1}, filling in the missing features, during which known features are also updated. Then, in 4th line of Algorithm \ref{aa1}, the known node attributes are reset, allowing them to participate in the next iteration. After completing the iterations, the final output \({\mathbf X}_{u}\), representing the reconstructed missing attributes, is denoted as \(\hat{\mathbf X}_{u}\). Both our attribute reconstruction and node classification tasks solely use \(\hat{\mathbf X}_{u}\) as input, with no other inputs involved. This method ensures that we assess the usability of the reconstructed features in isolation. For the node classification task, all methods use a two-layer MLP as the classifier, with an Adam optimizer set to a learning rate of 1e-2, a hidden channel dimension of 256, and a maximum training epoch of 1000, performing five-fold cross-validation \cite{rossi2022unreasonable}. The overall implementation process is illustrated in Figure \ref{figaa}.

\begin{table}[htbp]
  \centering \small
  \caption{Evaluation of the Accuracy (\%) of ARB and baseline methods on node classification. Best results are  \textcolor{blue}{\textbf{blue}}, second best results are \textcolor{darkgreen}{\textbf{green}}.}
        \renewcommand\arraystretch{1}
   \renewcommand\tabcolsep{3.5pt}
    \begin{tabular}{ccccccc}
    \toprule
    Dataset & SAT   & SVGA  & ITR   & FP    & PCFI  & ARB \\
    \midrule
    Cora  & 76.44  & 78.70  & 81.43  & {84.37}  & \textcolor{darkgreen}{\textbf{84.64}}  & \textcolor{blue}{\textbf{85.78}} \\
    CiteSeer & 60.10  & 62.33  & \textcolor{darkgreen}{\textbf{67.15}}  & 66.21  & 66.83  & \textcolor{blue}{\textbf{67.20}} \\
    PubMed & 46.18 & 62.27 & {72.36} & 80.62  & \textcolor{darkgreen}{\textbf{80.73}}  & \textcolor{blue}{\textbf{82.21}} \\
    Computers & 74.10  & 72.56  & {83.88}  & 83.71  & \textcolor{darkgreen}{\textbf{84.02}}  & \textcolor{blue}{\textbf{86.08}} \\
    Photo & 87.62  & 88.55  & \textcolor{darkgreen}{\textbf{90.75}}  & 87.41  & 89.56  & \textcolor{blue}{\textbf{91.74}} \\
    CS    & 76.72 & 82.93 & {85.67} & 89.30  & \textcolor{darkgreen}{\textbf{89.60}}  & \textcolor{blue}{\textbf{91.54}} \\
    Ogbn-Arxiv & 22.44  & 21.59  & 24.72  & \textcolor{darkgreen}{\textbf{51.50}}  & 51.21  & \textcolor{blue}{\textbf{52.48}} \\
    Ogbn-Products & 27.61  & 27.30  & 28.12  & {75.01}  & \textcolor{darkgreen}{\textbf{76.23}}  & \textcolor{blue}{\textbf{81.28}} \\
    \midrule
     \textit{Average} & 58.90  & 62.03  & 66.76  & {77.27}  & \textcolor{darkgreen}{\textbf{77.85}}  & \textcolor{blue}{\textbf{79.79}} \\
    \bottomrule
    \end{tabular}%
  \label{tab3}%
\end{table}%

\begin{table}[htbp]\small
  \centering
  \caption{Semi-supervised node classification accuracy (\%) at various missing rates. Best results are  \textcolor{blue}{\textbf{blue}}.}
    \renewcommand\arraystretch{0.8}
   \renewcommand\tabcolsep{3.5pt}
    \begin{tabular}{ccccccc}
    \toprule
    \multicolumn{1}{c}{\multirow{2}[4]{*}{Dataset}} & \multicolumn{3}{c}{50\%} & \multicolumn{3}{c}{90\%} \\
\cmidrule{2-7}          & FP    & PCFI  & ARB   & FP    & PCFI  & ARB \\
    \midrule
    Cora  & 80.71 & 80.51 & 81.38 & 79.09 & 78.84 & 80.20 \\
    CiteSeer & 65.58& 68.32 & 67.55 & 66.10 & 66.13 & 66.77 \\
    PubMed & 74.87 & 75.12 & {75.97} & 74.28 & 74.06 & {75.08} \\
    Photo & 91.07 & 90.38 & 91.43 & {88.67} & 88.32 & 90.01 \\
    Computers & {83.81} & 81.76 & {84.07} & {81.21} & {81.12} & {81.78} \\
    \midrule
    \textit{Average} & 79.21 & 79.22 & \color{blue}{\textbf{80.08}} & 77.87 & 77.69 & \color{blue}{\textbf{78.77}} \\
    \midrule
    \multicolumn{1}{c}{\multirow{2}[4]{*}{Dataset}} & \multicolumn{3}{c}{99\%} & \multicolumn{3}{c}{99.5\%} \\
\cmidrule{2-7}          & FP    & PCFI  & ARB   & FP    & PCFI  & ARB \\
    \midrule
    Cora  & 77.73 & 77.87 & 78.62 & 76.81 & 76.91 & 77.12 \\
    CiteSeer & 66.06 & 65.26 & 66.61 & 64.10 & 65.45 & 66.90 \\
    PubMed & 72.09 & 73.42 & {73.46} & 72.12 & 72.20 & 72.99 \\
    Photo & {88.13} & 87.98 & 88.44 & 87.29 & 87.80 & 88.09 \\
    Computers & {79.37} & {79.75} & 79.78 & 78.06 &  79.51 & 79.05 \\
    \midrule
    \textit{Average} & 76.68 & 76.85 & \color{blue}{\textbf{77.38}} & 75.67 & 76.37 & \color{blue}{\textbf{76.83}} \\
    \bottomrule
    \end{tabular}%
  \label{taba3}%
\end{table}%

\subsection{Attribute Reconstruction Results (\textbf{Q1})}
To evaluate the quality of attribute reconstruction, the similarity probability between the reconstructed attributes and the true attributes at each dimension is the main target. To achieve this, we follow the experimental setup of \cite{yoo2022accurate}, using Recall@k and nDCG@k as metrics to assess binary feature datasets with k set to \{10, 20, 50\}. For continuous feature datasets, we use RMSE and CORR as metrics. The results of the attribute reconstruction are summarized in Tables \ref{tab_2} and \ref{tab_21}. Our method consistently outperforms the baseline method, with an average improvement of \textbf{2.87\%} and \textbf{9.96\%} over the second-best method, respectively.

Further analysis of comparison shows that GNN based methods, such as SVGA, ITR and MATE, achieve competitive results on binary feature datasets compared to propagation-based baselines like FP and PCFI. However, in continuous feature datasets, propagation methods demonstrate a stronger advantage. This indicates that feature propagation-based methods have superior capabilities in complex attribute reconstruction. Additionally, since these methods do not provide effective solutions for the cold start problem, their performance remains consistently suboptimal compared to ARB.

Thus, in response to Q1, our conclusion is: \textbf{ARB exhibits strong performance in reconstructing features for attribute-missing graphs, effectively recovering missing attributes with high accuracy across various datasets.}

\subsection{Node Classification Results (\textbf{Q2})}
In our node classification experiments, our primary focus is to validate the effectiveness of the reconstructed features for downstream tasks. Therefore, we only use the reconstructed attribute features $\hat{\mathbf X}_{u}$ of the unknown nodes $\mathcal{V}_{u}$ for five-fold cross-validation, a choice influenced by methods like SAT \cite{chen2022learning} and SVGA \cite{yoo2022accurate}. We input the features that achieve the highest Recall@10 and CORR metrics in attribute reconstruction into a linear classifier for five-fold cross validation. The results of the node classification are summarized in Table \ref{tab3}. Our method consistently outperforms the baseline method, with an average improvement of \textbf{2.49\%} over the second-best method.

Further analysis reveals that while deep generative methods like SAT and SVGA demonstrate certain advantages in attribute reconstruction, they underperform in downstream node classification tasks, particularly on continuous feature datasets. Notably, a key advantage of propagation-based methods like FP and PCFI is their independence from gradient descent, enabling efficient training on CPU \cite{rossi2022unreasonable}. This eliminates the need for memory-intensive graph partitioning and batch processing, providing a natural advantage on large-scale datasets like Ogbn-Arxiv and Ogbn-Products. Moreover, ARB outperforms other propagation-based algorithms, demonstrating its strong reconstructed attributes in downstream tasks. 

However, we note that the experimental setup differs from those used in FP \cite{rossi2022unreasonable} and PCFI \cite{um2023confidence}, where semi-supervised node classification includes all nodes $\mathcal{V}$ with known attributes. Additionally, these methods follow the standard dataset splits and settings from PyG \footnote{https://pyg.org/} and OGBN \footnote{https://ogb.stanford.edu/docs/nodeprop/} and achieve competitive results even under extreme missing rates (99\% and 99.5\%). To ensure a fair comparison and to assess ARB's effectiveness in semi-supervised node classification, we conducted additional experiments under these conditions. As shown in Table \ref{taba3}, ARB outperforms other methods, demonstrating superior performance across both regular and extreme missing conditions.

Thus, in response to Q2, our conclusion is: \textbf{ARB demonstrates strong adaptability in downstream tasks after attribute reconstruction, outperforming baseline methods, particularly in large-scale datasets.}

\begin{figure}[h]
\centering
\subfloat[Early stop on Cora]{
\includegraphics[width=0.22\textwidth]{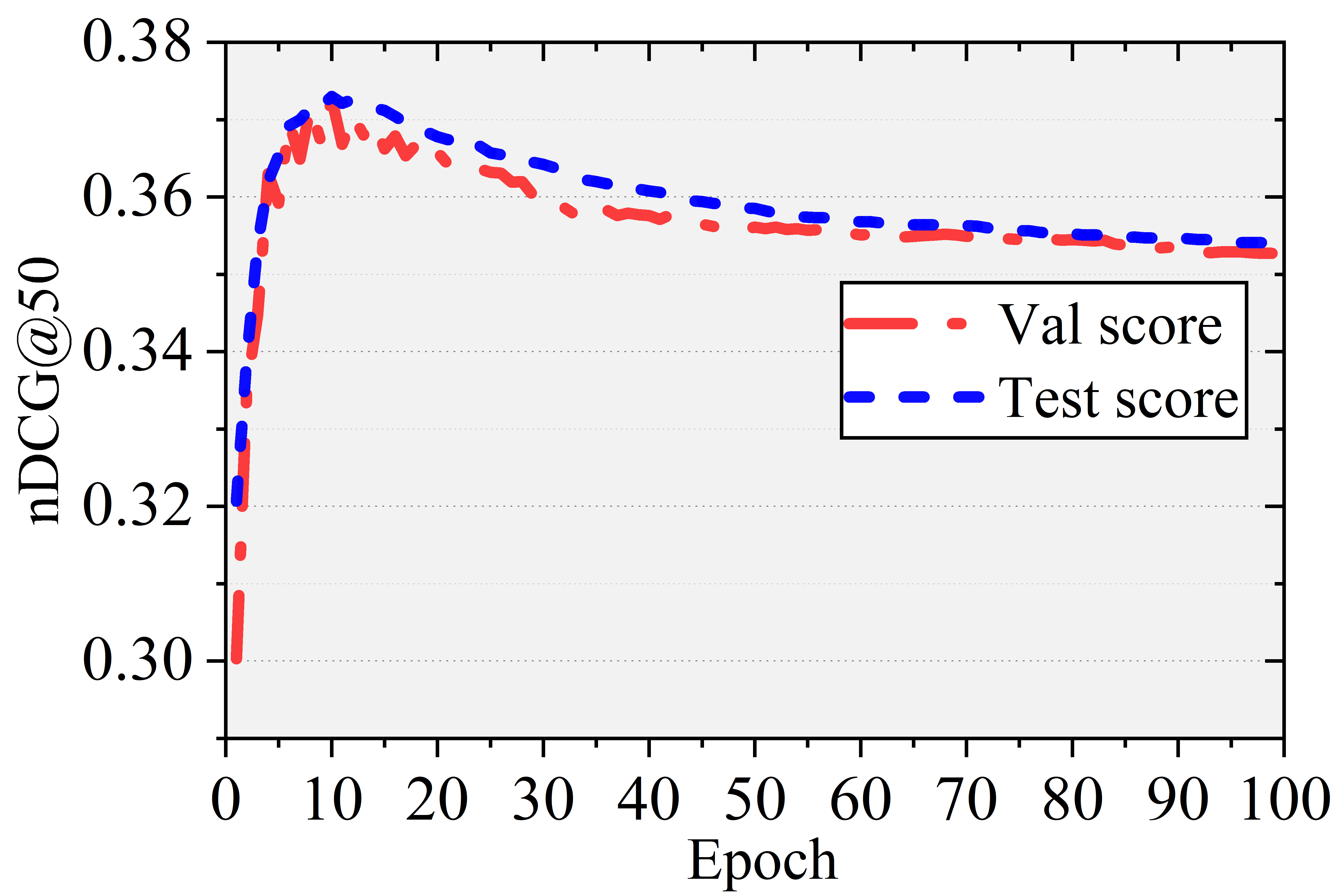}
}
\hspace{0.1cm}
\centering
\subfloat[\small Early stop on PubMed]{
\includegraphics[width=0.22\textwidth]{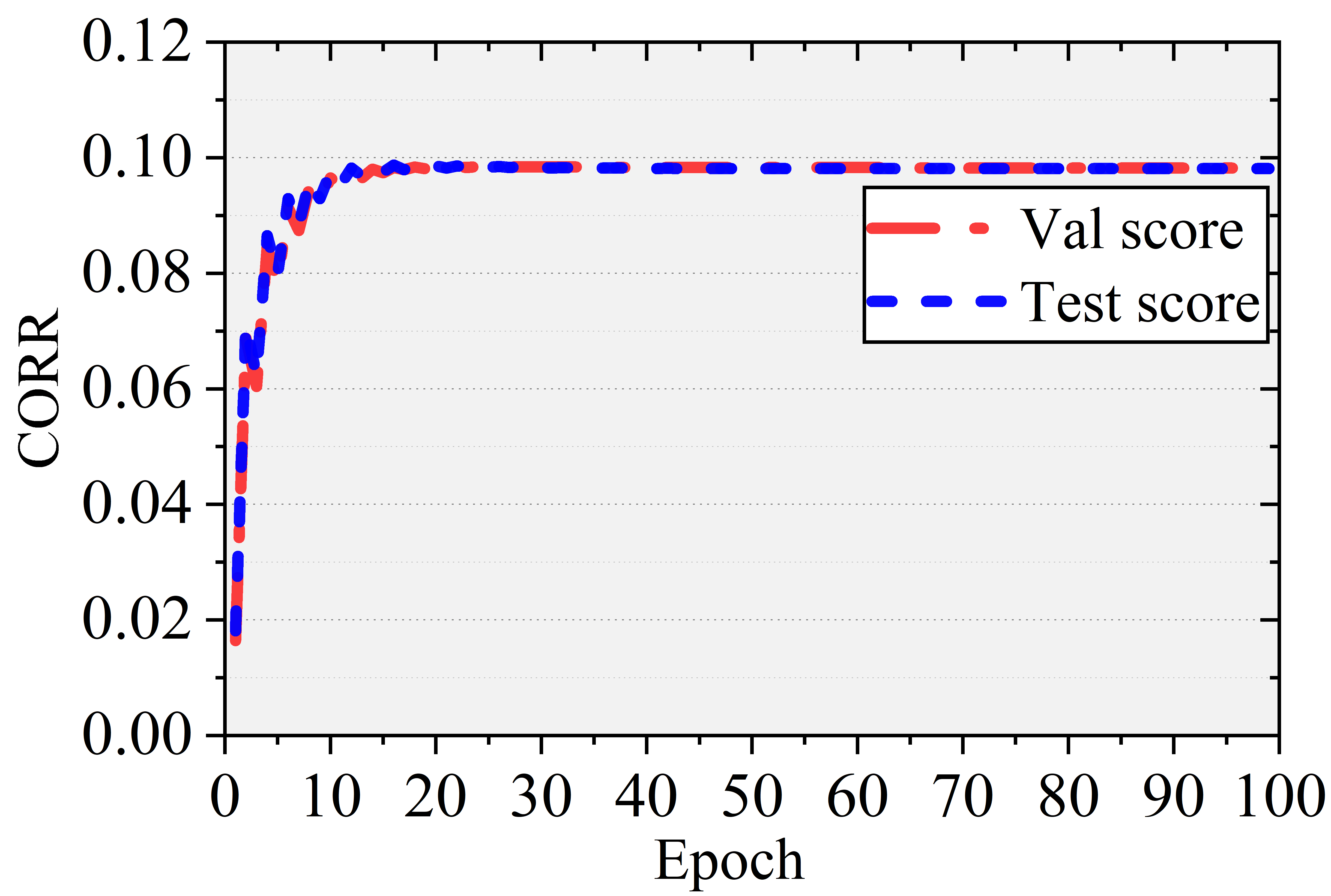}
}

\centering
\subfloat[\small Convergence speed comparison of baselines on Computers]{
\includegraphics[width=0.22\textwidth]{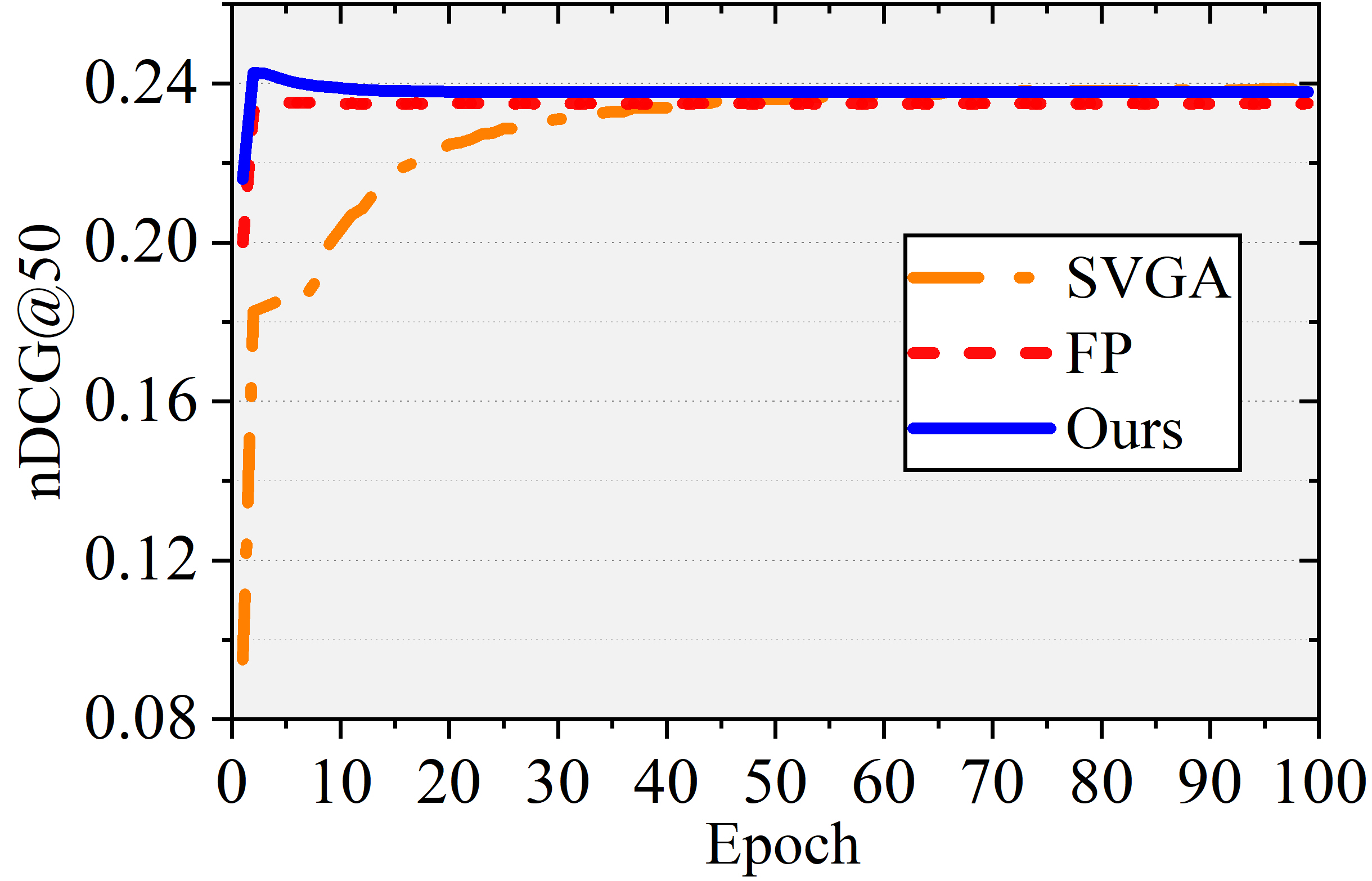}
}
\hspace{0.1cm}
\centering
\subfloat[\small Convergence speed comparison on PubMed]{
\includegraphics[width=0.22\textwidth]{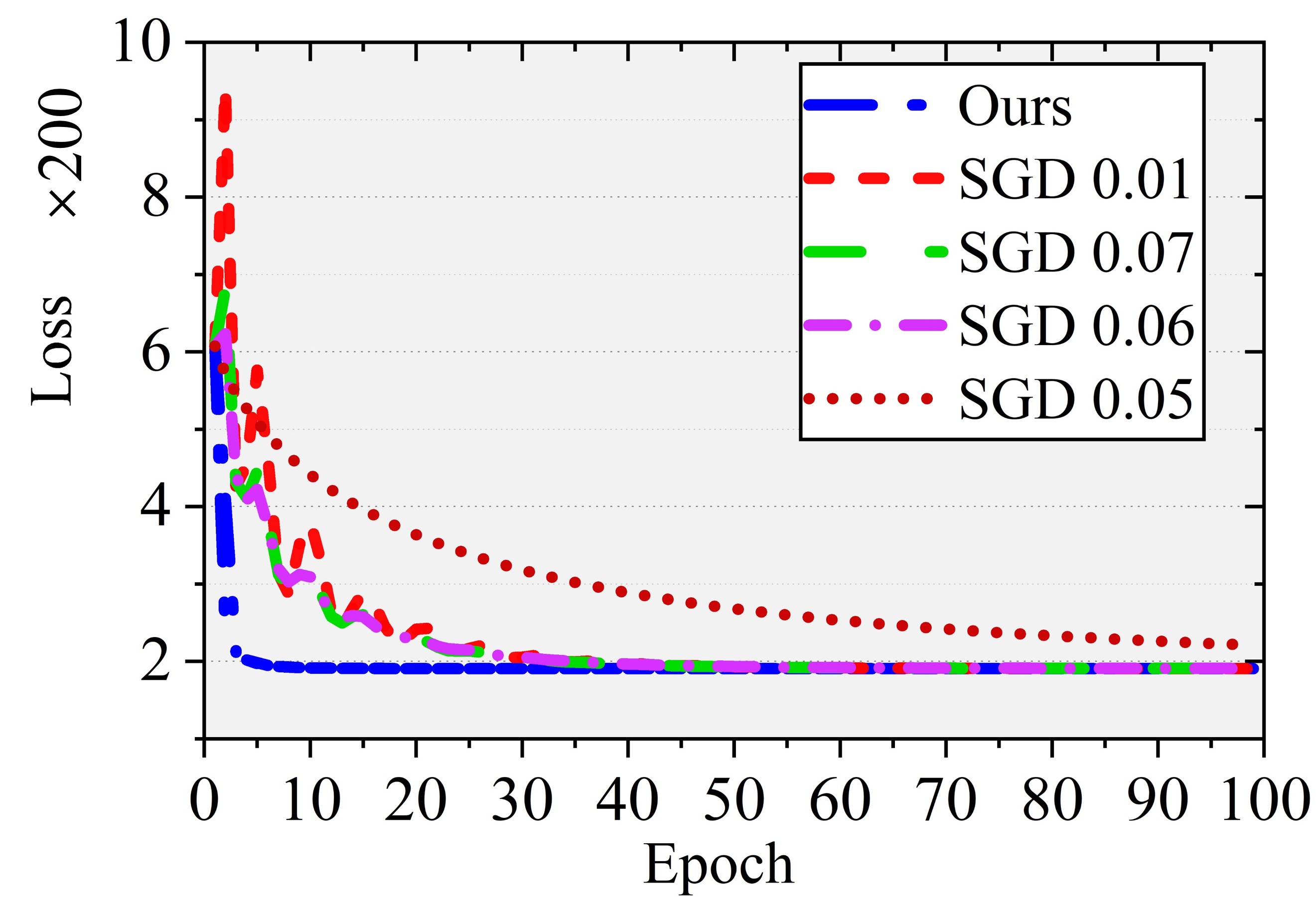}
}
\caption{Training process and convergence speed.}
\label{fig_2}
\end{figure}

\begin{table*}[htbp]
  \centering
\small
  \caption{Evaluation of ARB and baseline methods on binary features for attribute reconstruction. Best results are indicated in \textcolor{blue}{\textbf{blue}}. NA means not available.}
    \renewcommand\arraystretch{1}
   \renewcommand\tabcolsep{3.8pt}
    \begin{tabular}{ccccccccccccccc}
    \toprule
    \multirow{1}[2]{*}{Metric} & \multirow{1}[2]{*}{Method} & \multirow{1}[2]{*}{Node Type}& \multicolumn{3}{c}{Cora} & \multicolumn{3}{c}{CiteSeer} & \multicolumn{3}{c}{Computers} & \multicolumn{3}{c}{Photo} \\
\cline{4-15}  &   & &  @10  &  @20  &  @50  &  @10  &  @20  &  @50  &  @10  &  @20  &  @50  &  @10  &  @20  &  @50  \\
    \hline
    \multicolumn{1}{c}{\multirow{6}[6]{*}{{Recall}}} & w/o BC\&VE    & isolated & NA    & NA    & NA    & 0.0000  & 0.0052  & 0.0151  & 0.0120  & 0.0259  & 0.0609  & 0.0125  & 0.0285  & 0.0653  \\
          & ARB   & isolated & NA    & NA    & NA    & \color{blue}{\textbf{0.0457}} & \color{blue}{\textbf{0.0916}} & \color{blue}{\textbf{0.1701}} & \color{blue}{\textbf{0.0308}} & \color{blue}{\textbf{0.0616}} & \color{blue}{\textbf{0.1441}} & \color{blue}{\textbf{0.0332}} & \color{blue}{\textbf{0.0611}} & \color{blue}{\textbf{0.1390}} \\
\cmidrule{2-15}          & w/o BC\&VE    & low degree & 0.1504  & 0.2075  & 0.3117  & 0.0777  & 0.1286  & 0.2171  & 0.0463  & 0.0802  & 0.1611  & 0.0382  & 0.0696  & 0.1487  \\
          & ARB   & low degree & \color{blue}{\textbf{0.1757}} & \color{blue}{\textbf{0.2475}} & \color{blue}{\textbf{0.3735}} & \color{blue}{\textbf{0.0981}} & \color{blue}{\textbf{0.1569}} & \color{blue}{\textbf{0.2710}} & \color{blue}{\textbf{0.0497}} & \color{blue}{\textbf{0.0841}} & \color{blue}{\textbf{0.1715}} & \color{blue}{\textbf{0.0481}} & \color{blue}{\textbf{0.0855}} & \color{blue}{\textbf{0.1739}} \\
\cmidrule{2-15}          & w/o BC\&VE    & other & 0.1839  & 0.2482  & 0.3708  & 0.1250  & 0.1870  & 0.2990  & 0.0430  & 0.0750  & 0.1566  & 0.0436  & 0.0775  & 0.1629  \\
          & ARB   & other & \color{blue}{\textbf{0.1903}} & \color{blue}{\textbf{0.2771}} & \color{blue}{\textbf{0.4055}} & \color{blue}{\textbf{0.1257}} & \color{blue}{\textbf{0.1915}} & \color{blue}{\textbf{0.3175}} & \color{blue}{\textbf{0.0442}} & \color{blue}{\textbf{0.0771}} & \color{blue}{\textbf{0.1604}} & \color{blue}{\textbf{0.0441}} & \color{blue}{\textbf{0.0790}} & \color{blue}{\textbf{0.1657}} \\
    \midrule
    \multicolumn{1}{c}{\multirow{6}[6]{*}{{nDCG}}} & w/o BC\&VE    & isolated & NA    & NA    & NA    & 0.0000  & 0.0107  & 0.0108  & 0.0325  & 0.0462  & 0.0795  & 0.0358  & 0.0534  & 0.0895  \\
          & ARB   & isolated & NA    & NA    & NA    & \color{blue}{\textbf{0.0613}} & \color{blue}{\textbf{0.1001}} & \color{blue}{\textbf{0.1525}} & \color{blue}{\textbf{0.0687}} & \color{blue}{\textbf{0.1057}} & \color{blue}{\textbf{0.1848}} & \color{blue}{\textbf{0.0766}} & \color{blue}{\textbf{0.1128}} & \color{blue}{\textbf{0.1934}} \\
\cmidrule{2-15}          & w/o BC\&VE    & low degree & 0.2086  & 0.2457  & 0.3029  & 0.1271  & 0.1708  & 0.2306  & 0.1059  & 0.1469  & 0.2255  & 0.0879  & 0.1272  & 0.2037  \\
          & ARB   & low degree & \color{blue}{\textbf{0.2451}} & \color{blue}{\textbf{0.2928}} & \color{blue}{\textbf{0.3593}} & \color{blue}{\textbf{0.1645}} & \color{blue}{\textbf{0.2135}} & \color{blue}{\textbf{0.2888}} & \color{blue}{\textbf{0.1109}} & \color{blue}{\textbf{0.1529}} & \color{blue}{\textbf{0.2375}} & \color{blue}{\textbf{0.1118}} & \color{blue}{\textbf{0.1581}} & \color{blue}{\textbf{0.2453}} \\
\cmidrule{2-15}          & w/o BC\&VE   & other & 0.2623  & 0.3061  & 0.3720  & 0.2141  & 0.2674  & 0.3438  & 0.1073  & 0.1506  & 0.2387  & 0.1069  & 0.1516  & 0.2417  \\
          & ARB   & other & \color{blue}{\textbf{0.2695}} & \color{blue}{\textbf{0.3272}} & \color{blue}{\textbf{0.3968}} & \color{blue}{\textbf{0.2177}} & \color{blue}{\textbf{0.2742}} & \color{blue}{\textbf{0.3592}} & \color{blue}{\textbf{0.1098}} & \color{blue}{\textbf{0.1562}} & \color{blue}{\textbf{0.2460}} & \color{blue}{\textbf{0.1081}} & \color{blue}{\textbf{0.1548}} & \color{blue}{\textbf{0.2497}} \\
    \bottomrule
    \end{tabular}%
  \label{tabadd}%
\end{table*}%

\subsection{Convergence Speed to Verify Cold Start (\textbf{Q3})}
Judging ARB's performance in mitigating the cold start problem could be done by evaluating its early stopping and convergence speed capabilities.

As shown in Figure \ref{fig_2}(a), in weakly connected graphs like Cora, ARB reaches the optimal solution in fewer epochs (Epoch $\approx$ 10), demonstrating faster attribute reconstruction. Similarly, in strongly connected graphs like PubMed (Figure \ref{fig_2}(b)), ARB adapts and approaches the optimal solution by Epoch $\approx$ 15, highlighting its efficiency in resolving the cold start problem with minimal training.

During the iterative process, ARB converges faster than deep generative methods. As shown in Figure \ref{fig_2}(c), it outperforms SVGA and FP in both convergence speed and reconstruction accuracy, achieving satisfactory results early (Epoch=2) and enabling early stopping. This rapid convergence reduces overall computation time, making ARB more efficient for large-scale and time-sensitive applications. Similar to SGD-based methods, ARB initializes unknown node attributes to zero while preserving known ones. However, ARB's dynamic propagation method significantly outperforms SGD in terms of convergence speed, as shown in Figure \ref{fig_2}(d), by adjusting the boundary of known nodes and propagating information without the need for backpropagation.

Furthermore, we conduct a separate evaluation of isolated nodes and low-degree nodes to verify the effectiveness of the ARB method in the attribute reconstruction task. Table \ref{tabadd} shows a significant performance improvement when comparing the FP and ARB models, particularly for isolated nodes and low-degree nodes. The proposed ARB method substantially enhances the attribute reconstruction performance for these nodes, effectively addressing the cold-start problem. This further validates the effectiveness and practicality of our method.

Thus, in response to Q3, our conclusion is: \textbf{ARB is able to effectively tackle the cold start problem and convergence difficulty, ensuring more stable and rapid convergence.}

\begin{table*}[t]
\centering
\small
  \caption{Ablation experiments for attribute reconstruction. Best results are  \textcolor{blue}{\textbf{blue}}.}
      \renewcommand\arraystretch{1.2}
   \renewcommand\tabcolsep{3.5pt}
    \begin{tabular}{cccccccccccccc}
    \toprule
    \multirow{1}[4]{*}{Metric} & \multicolumn{1}{c}{\multirow{1}[4]{*}{Method}} & \multicolumn{3}{c}{Cora} & \multicolumn{3}{c}{CiteSeer} & \multicolumn{3}{c}{Computers} & \multicolumn{3}{c}{Photo} \\
\cline{3-14}      &   & \multicolumn{1}{c}{@10} & \multicolumn{1}{c}{@20} & \multicolumn{1}{c}{@50} & \multicolumn{1}{c}{@10} & \multicolumn{1}{c}{@20} & \multicolumn{1}{c}{@50} & \multicolumn{1}{c}{@10} & \multicolumn{1}{c}{@20} & \multicolumn{1}{c}{@50} & \multicolumn{1}{c}{@10} & \multicolumn{1}{c}{@20} & \multicolumn{1}{c}{@50} \\
    \hline
    \multirow{3}[2]{*}{Recall} 
    & w/o BC & 0.1555  & 0.2199  & 0.3303  & 0.0821  & 0.1339  & 0.2266  & 0.0430  & 0.0750  & 0.1557  & 0.0436  & 0.0773  & 0.1622  \\
        & w/o VE & 0.1771  & 0.2488  & 0.3693  & 0.0927  & 0.1482  & 0.2470  & 0.0442  & 0.0770  & 0.1595  & 0.0447  & 0.0793  & 0.1655  \\
      & ARB & \textcolor{blue}{\textbf{0.1856}}  & \textcolor{blue}{\textbf{0.2599}}  & \textcolor{blue}{\textbf{0.3851}} & \textcolor{blue}{\textbf{0.1046}} & \textcolor{blue}{\textbf{0.1643}} & \textcolor{blue}{\textbf{0.2823}} & \textcolor{blue}{\textbf{0.0449}}  & \textcolor{blue}{\textbf{0.0784}} & \textcolor{blue}{\textbf{0.1627}} & \textcolor{blue}{\textbf{0.0455}}  & \textcolor{blue}{\textbf{0.0804}}  & \textcolor{blue}{\textbf{0.1681}}  \\
    \hline
    \multirow{4}[2]{*}{nDCG} & w/o BC & 0.2195  & 0.2637  & 0.3220  & 0.1373  & 0.1828  & 0.2444  & 0.1060  & 0.1488  & 0.2349  & 0.1064  & 0.1512  & 0.2411  \\
          & w/o VE & 0.2491  & 0.2968  & 0.3611  & 0.1576  & 0.2059  & 0.2712  & 0.1086  & 0.1522  & 0.2402  & 0.1087  & 0.1544  & 0.2458  \\
      & ARB & \textcolor{blue}{\textbf{0.2594}} & \textcolor{blue}{\textbf{0.3043}} & \textcolor{blue}{\textbf{0.3749}} & \textcolor{blue}{\textbf{0.1768}} & \textcolor{blue}{\textbf{0.2267}} & \textcolor{blue}{\textbf{0.3043}} & \textcolor{blue}{\textbf{0.1107}}  & \textcolor{blue}{\textbf{0.1553}}  & \textcolor{blue}{\textbf{0.2450}}  & \textcolor{blue}{\textbf{0.1104}}  & \textcolor{blue}{\textbf{0.1565}}  & \textcolor{blue}{\textbf{0.2493}}  \\
    \bottomrule
    \end{tabular}%
  \label{tab_6}%
\end{table*}

\begin{table}[htbp]
    \small
      \centering
  \caption{Ablation experiments for node classification. Best results are  \textcolor{blue}{\textbf{blue}}.}
        \renewcommand\arraystretch{0.8}
   \renewcommand\tabcolsep{4pt}
    \begin{tabular}{ccccc}
    \toprule
    Method & Cora  & CiteSeer & PubMed & Computers \\
    \midrule
    w/o BC & 84.42  & 66.89  & 81.61  & 84.65  \\
    w/o VE & 84.86  & 66.54  & 82.12  & 85.83  \\
    ARB   & \textcolor{blue}{\textbf{85.78}} & \textcolor{blue}{\textbf{67.20}} & \textcolor{blue}{\textbf{82.81}} & \textcolor{blue}{\textbf{86.08}} \\
    \midrule
    Method & Photo & CS    & Ogbn-Arxiv & Ogbn-Products \\
    \midrule
    w/o BC & 90.23  & 90.38  & 49.91  & 79.56  \\
    w/o VE & 90.56  & 89.57  & 52.01  & 80.44  \\
    ARB   & \textcolor{blue}{\textbf{91.74}} & \textcolor{blue}{\textbf{91.54}} & \textcolor{blue}{\textbf{52.48}} & \textcolor{blue}{\textbf{81.28}} \\
    \bottomrule
    \end{tabular}%
  \label{tab_61}%
\end{table}

\subsection{Ablation and Hyperparameters Experiments (\textbf{Q4})}
Tables \ref{tab_6} and \ref{tab_61} present the results of the ablation study for ARB. In the tables, ``w/o BC" denotes the removal of the new boundary condition mechanism in ARB, and ``w/o VE" indicates the removal of the virtual edge mechanism in ARB. The results show that ARB always achieves optimal performance across various scenarios, outdoing the other schemes in performance. Further analysis reveals that the new boundary conditions contribute more significantly to performance improvement compared to the virtual edges. While the virtual edge mechanism enhances global connectivity, it inevitably brings some noise. Therefore, to fully leverage its benefits, proper tuning of the parameters $\alpha$ and $\beta$ is necessary.

The virtual edges and new boundary conditions add a propagation channel but may also introduce noise. To address this, the hyperparameter \(\alpha\) regulates their influence weight. Additionally, the number of propagation layers \(l\) also impacts the results. Therefore, the three key hyperparameters—\(\alpha\), \(\beta\), and \(l\)—are critical to ARB's performance and require careful tuning.

Regarding hyperparameter tuning, we propose the Heuristic Hyperparameter Searcher, using nDCG or CORR as the target metric. Starting at \((\alpha, \beta) = (0.5, 0.5)\), neighboring points at distance \(d\) are evaluated, and the best-scoring point becomes the new center. The process repeats, halving \(d\) when no better points are found, until the stopping condition is met.

\begin{figure}[htbp]
\centering
\subfloat[Cora]{
\includegraphics[width=0.142\textwidth]{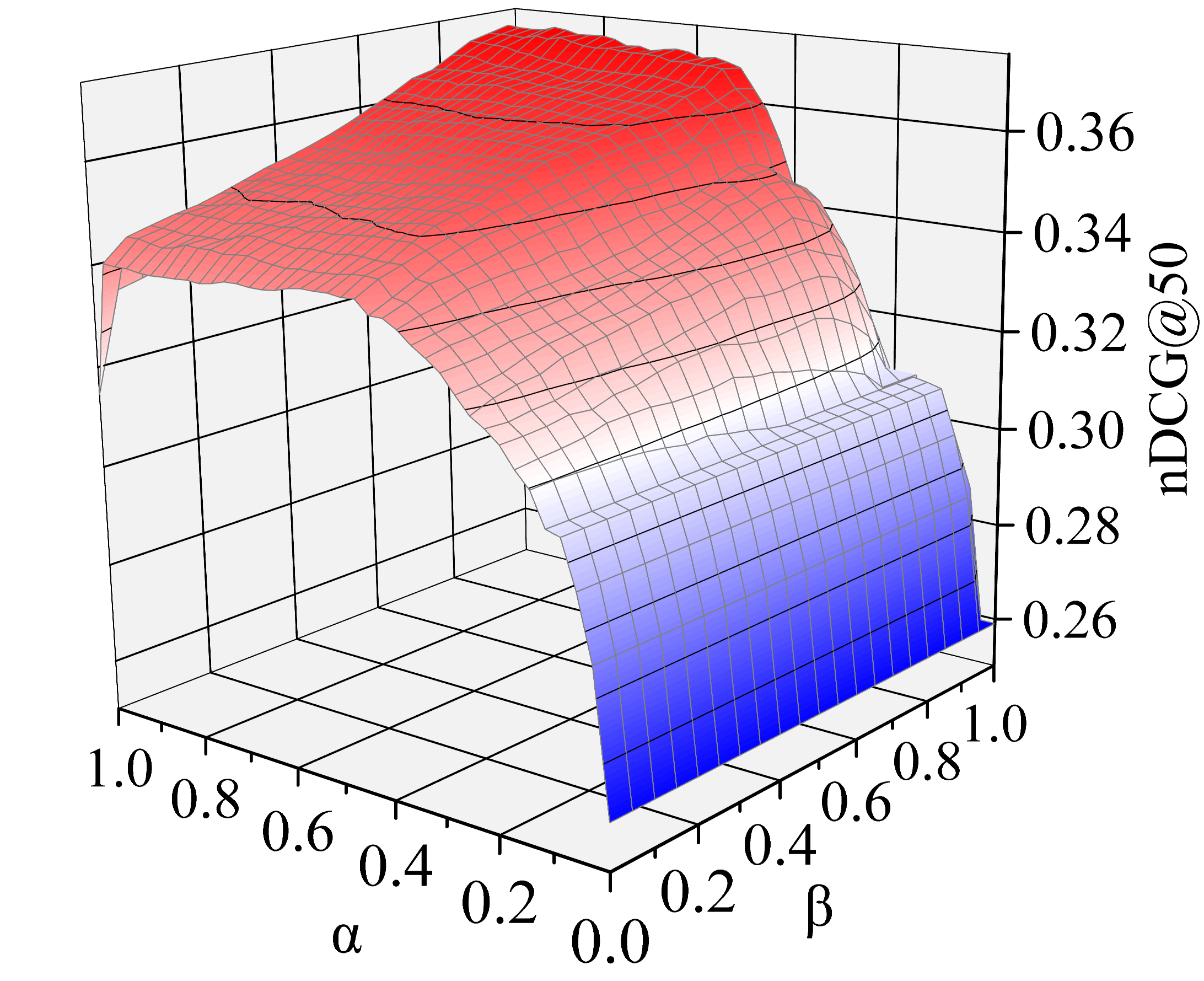}
}
\subfloat[CiteSeer]{
\includegraphics[width=0.142\textwidth]{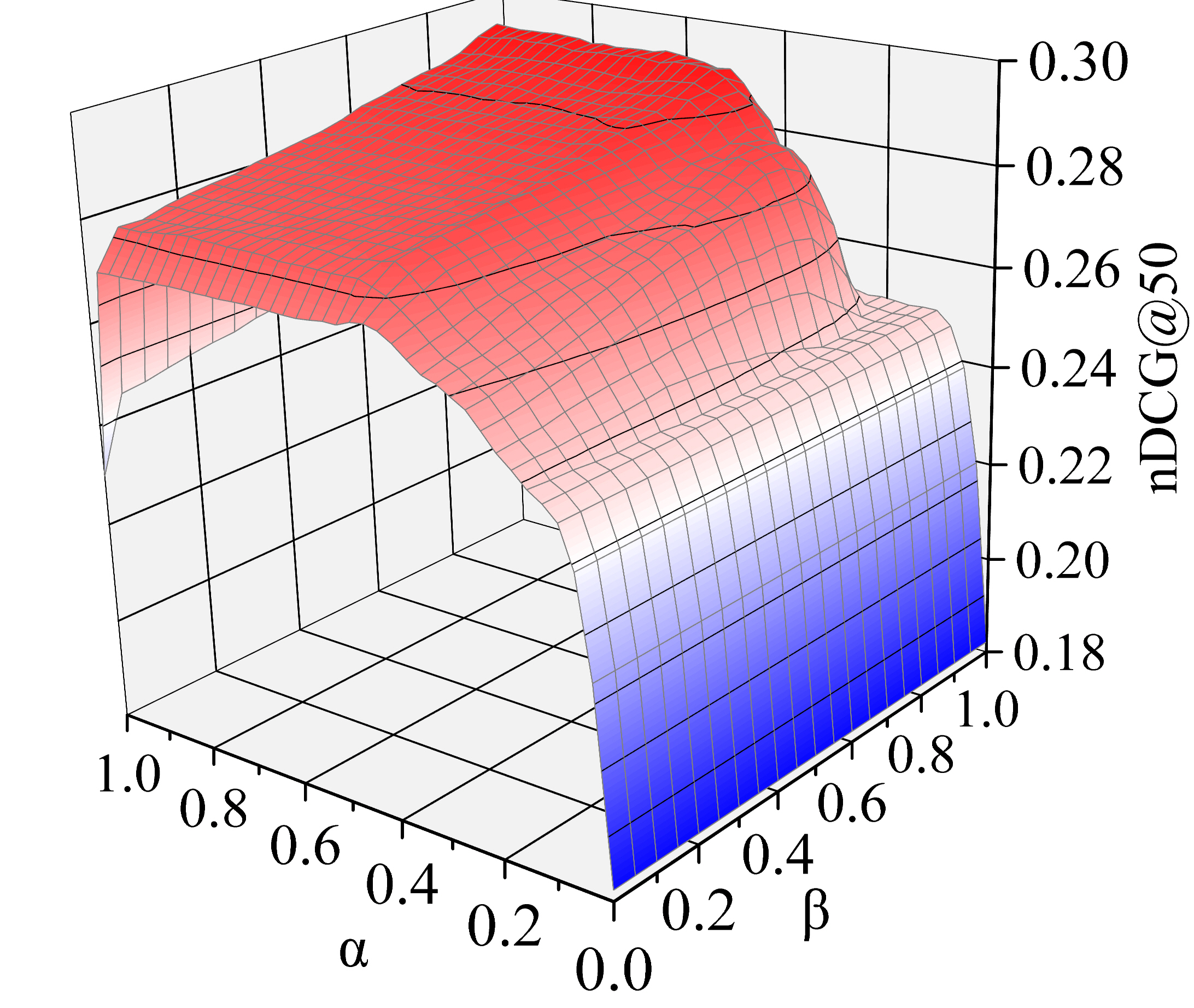}
}
\subfloat[Computers]{
\includegraphics[width=0.142\textwidth]{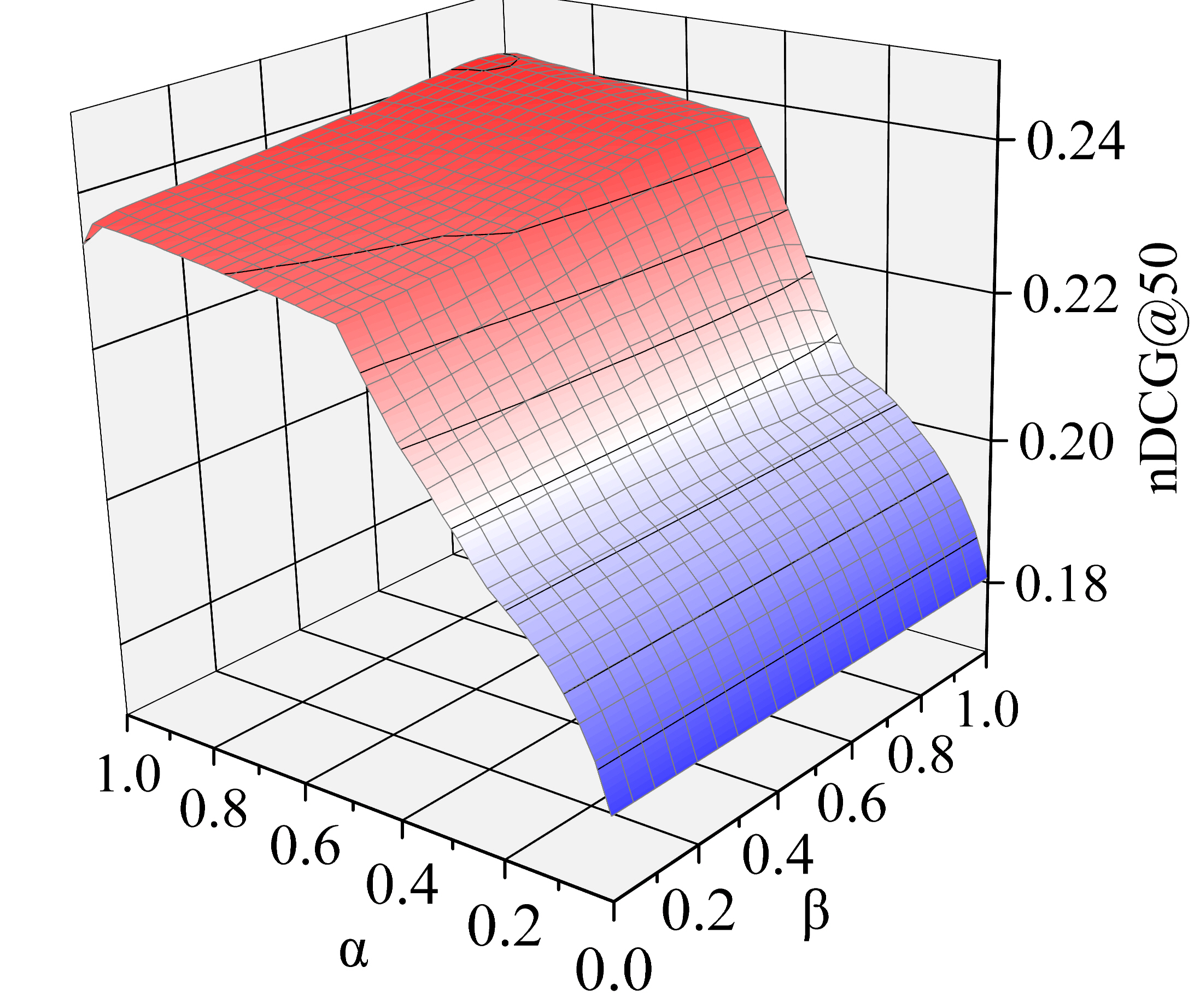}
}

\subfloat[Photo]{
\includegraphics[width=0.142\textwidth]{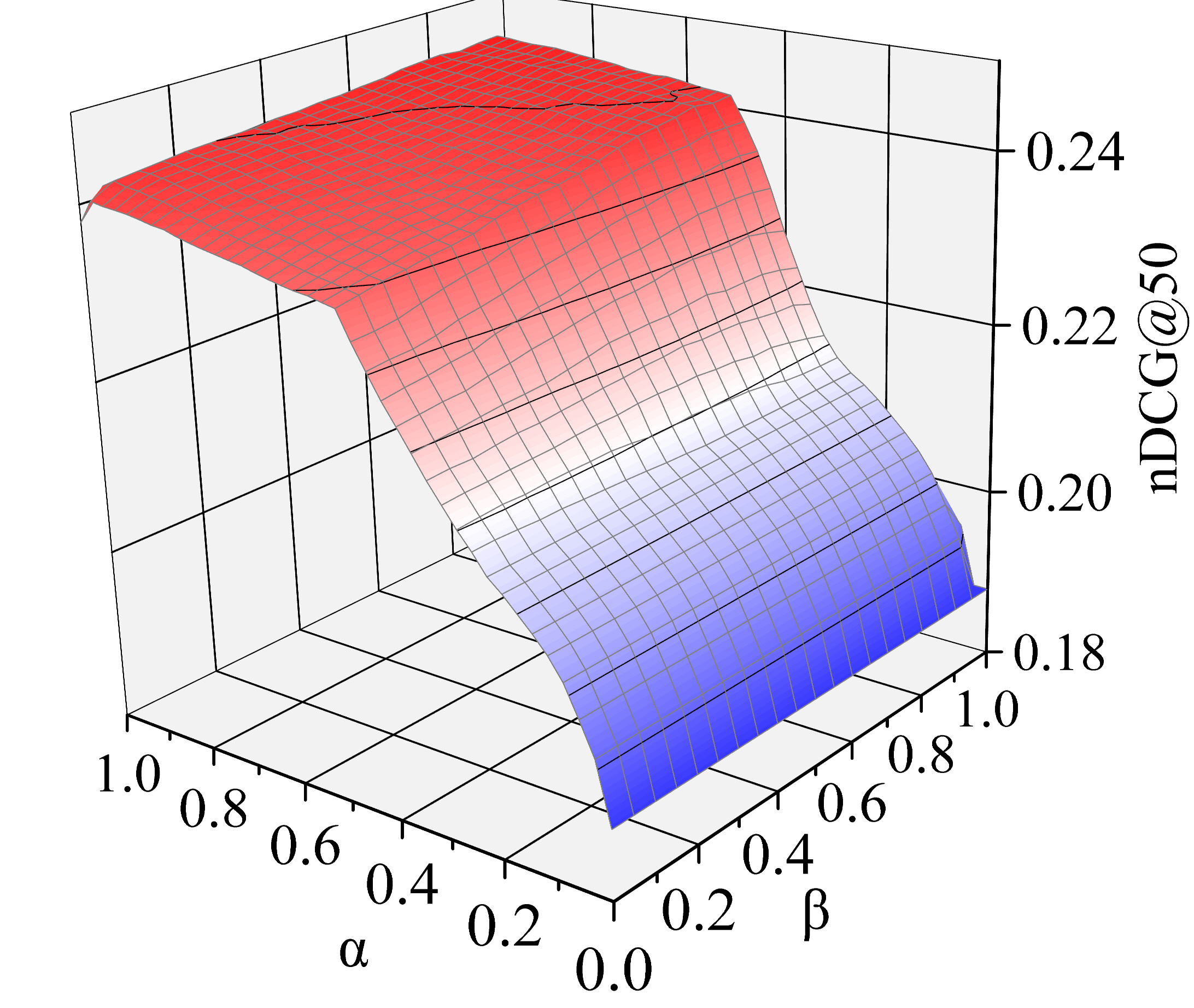}
}
\subfloat[PubMed]{
\includegraphics[width=0.142\textwidth]{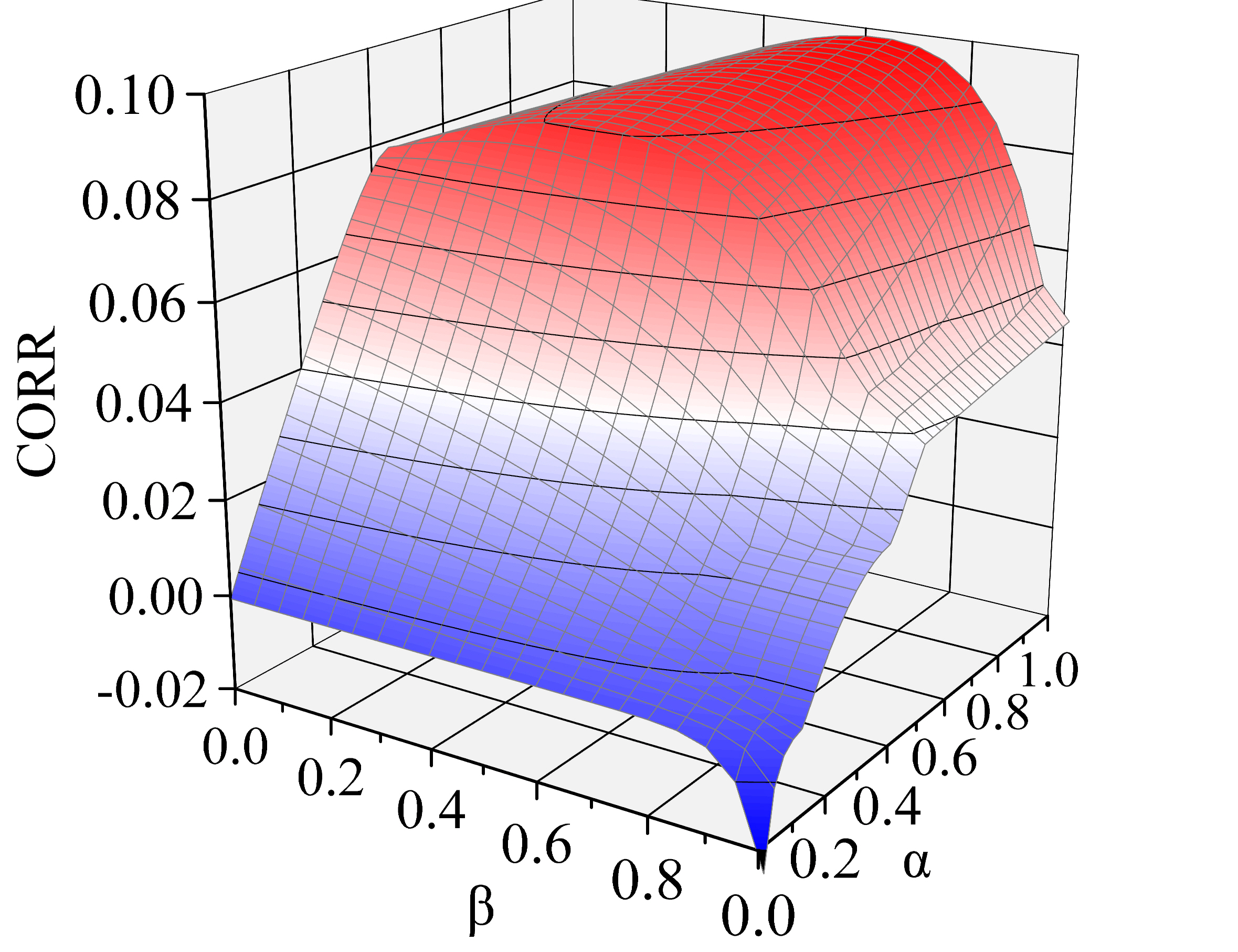}
}
\subfloat[CS]{
\includegraphics[width=0.142\textwidth]{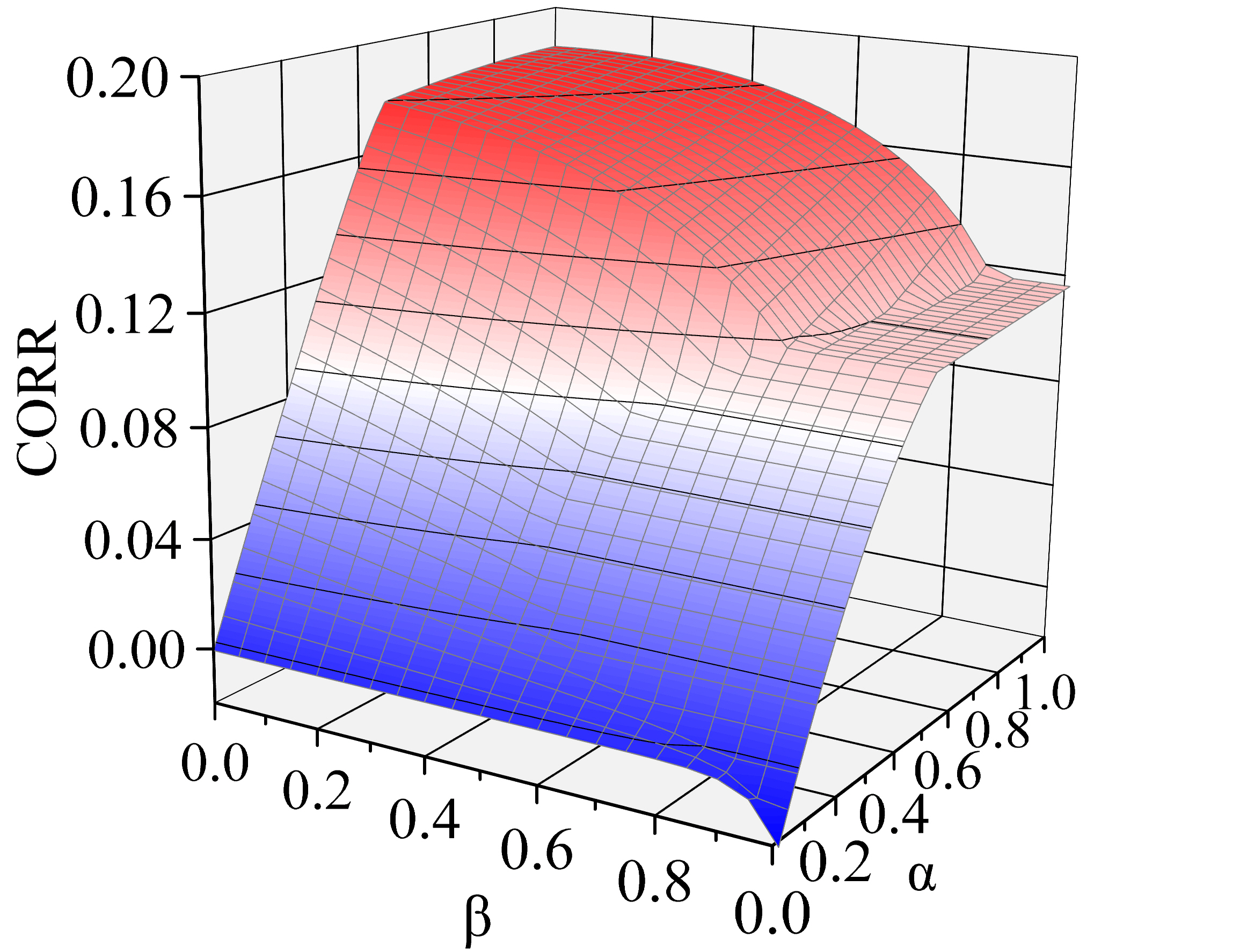}
}
\caption{Hyperparameter value $\alpha$ and $\beta$ validation.}
\label{fig_4}
\end{figure}
\begin{figure}[h]
\centering
\subfloat[Cora]{
\includegraphics[width=0.2\textwidth]{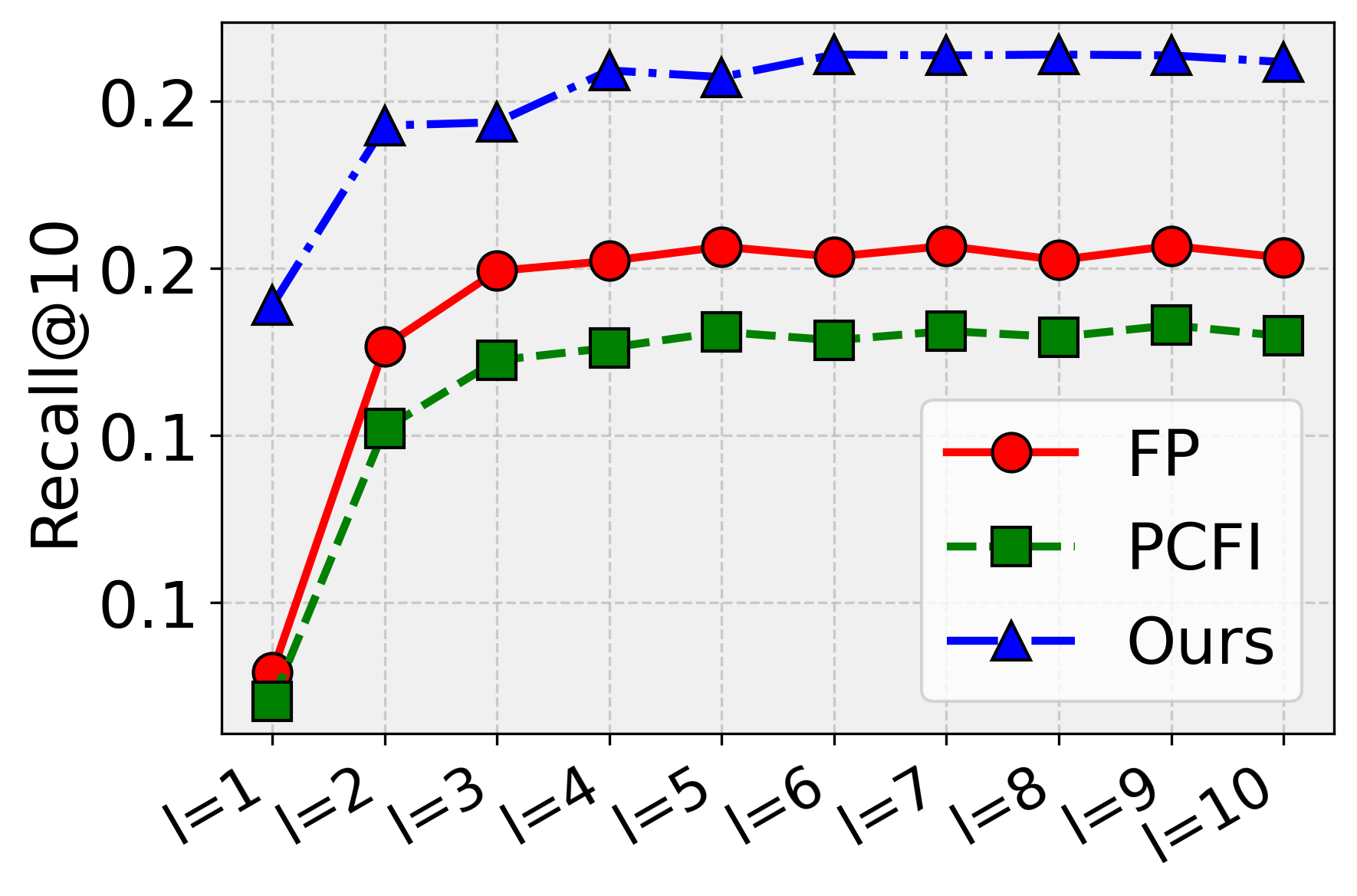}
}
\subfloat[CiteSeer]{
\includegraphics[width=0.2\textwidth]{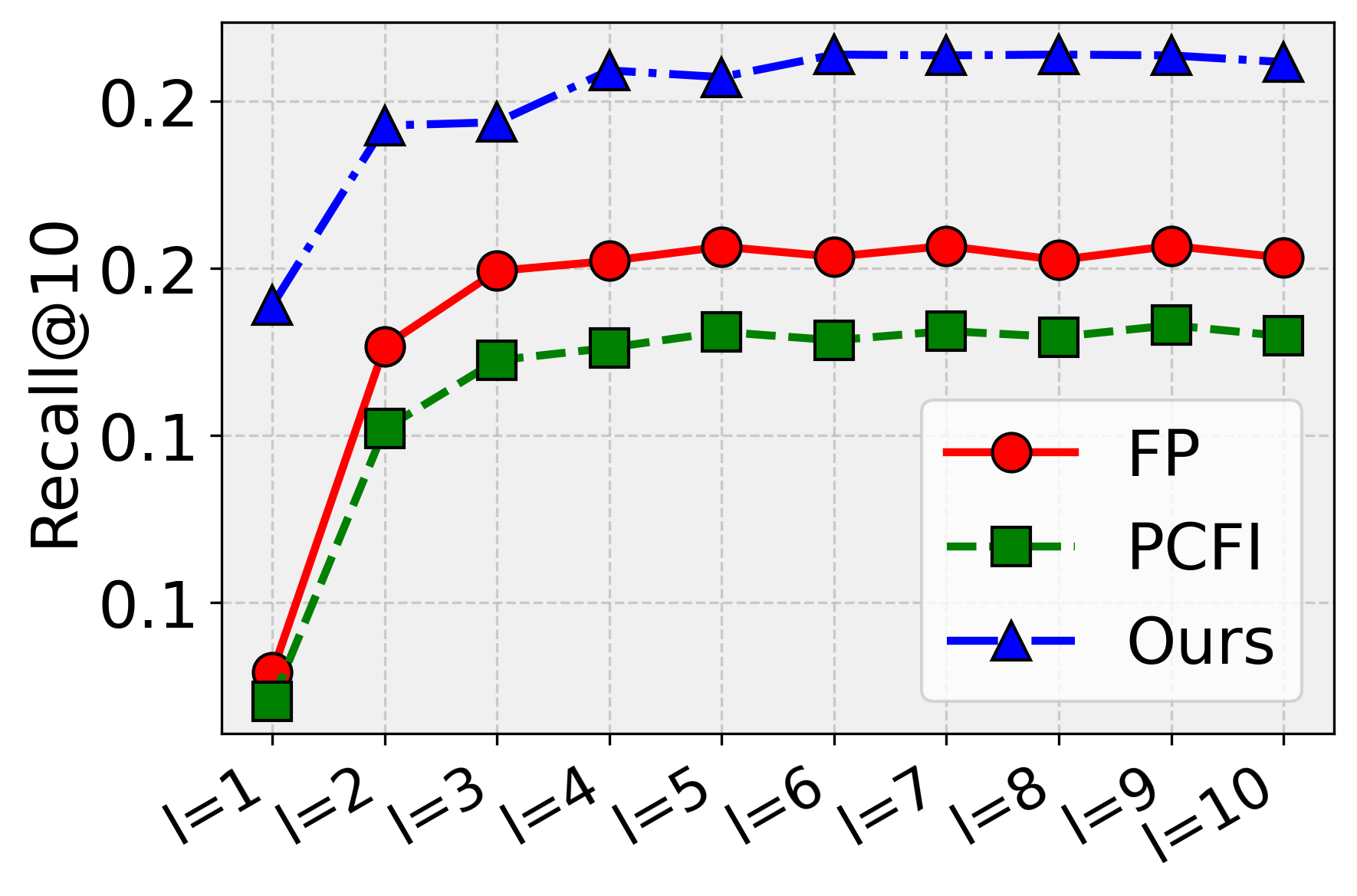}
}
\caption{Comparison of different propagation times $l$.}
\label{fig_a1}
\end{figure}

Concluded from Figure \ref{fig_4}, the optimal result suggests setting \(\alpha\) within the range of 0.9 to 1. \(\beta\) is crucial for the redefinition of boundary conditions in ARB, and must be tuned for each specific dataset to achieve optimal performance. 
At lower \(l\) values, the focus is on local neighborhood structures, which is crucial for nodes in small components with limited connectivity. As \(l\) increases, global propagation enriches the feature space with structural information from distant parts of the network. Figure \ref{fig_a1} shows that ARB effectively addresses propagation barriers and oversmoothing issues across depths from \(l = 1\) to \(l = 10\), significantly outperforming algorithms like FP and PCFI. In contrast, SVGA suffers from oversmoothing as \(l\) increases, leading to performance degradation, while ARB consistently maintains or improves performance with deeper propagation.

Thus, in response Q4, our conclusion is: \textbf{Each component of ARB has been validated and proven effective. Removing any component, including boundary conditions or virtual edges, leads to a drop in performance. The hyperparameters $\alpha$, $\beta$ and $l$ need to be adjusted for different datasets to achieve the best results.}

\subsection{Sensitivity Analysis on Missing Rates (\textbf{Q5})}
We conduct a series of attribute reconstruction experiments to validate the robustness of ARB, with missing attribute rates ranging from 40\% to 99\%, using Recall@10 as the primary evaluation metric.
As shown in Table \ref{tabm}, ARB demonstrates exceptional performance in attribute reconstruction tasks, particularly under high missing rates, confirming its robustness. Additionally, we observe that FP outperforms SVGA and ITR at higher missing rates, further highlighting the robustness of feature propagation-based methods. 

Therefore, for Question 5, our conclusion is: \textbf{ARB maintains robust performance across different missing rates, even when the missing rate is as high as 99\%.}
\begin{table}[htbp]\scriptsize
  \centering
  \caption{Comparison of methods under different missing rates (\%). MR stands for missing rate. Best results are \textcolor{blue}{\textbf{blue}}.} 
        \renewcommand\arraystretch{1.1}
   \renewcommand\tabcolsep{2.5pt}
    \begin{tabular}{ccccc|ccccc}
    \toprule
    Cora  & \multicolumn{4}{c}{Recall@10} & CiteSeer & \multicolumn{4}{c}{Recall@10} \\
    \midrule
    MR & SVGA  & ITR   & FP  & ARB  & MR & SVGA  & ITR   & FP  & ARB \\
    \midrule
    
    40\%   & 0.1876  & 0.1771  & 0.1675  & \textcolor{blue}{\textbf{0.1953}} & 40\%    & 0.1041  & 0.1059  & 0.1021  & \textcolor{blue}{\textbf{0.1184}} \\
    50\%   & 0.1804  & 0.1737  & 0.1632  & \textcolor{blue}{\textbf{0.1893}} & 50\%   & 0.0996  & 0.1016  & 0.0933  & \textcolor{blue}{\textbf{0.1094}} \\

    60\%   & 0.1718  & 0.1656  & 0.1620  & \textcolor{blue}{\textbf{0.1856}} & 60\%    & 0.0943  & 0.0972  & 0.0850  & \textcolor{blue}{\textbf{0.1046}} \\

    70\%   & 0.1650  & 0.1546  & 0.1607  & \textcolor{blue}{\textbf{0.1781}} & 70\%    & 0.0846  & 0.0863  & 0.0830  & \textcolor{blue}{\textbf{0.0983}} \\

    80\%   & 0.1586  & 0.1425  & 0.1582  & \textcolor{blue}{\textbf{0.1681}} & 80\%   & 0.0742  & 0.0740  & 0.0821  & \textcolor{blue}{\textbf{0.0897}} \\
    90\%   & 0.1376  & 0.1102  & 0.1487  & \textcolor{blue}{\textbf{0.1602}} & 90\%    & 0.0607  & 0.0589  & 0.0623  & \textcolor{blue}{\textbf{0.0844}} \\
    99\%   & 0.1121  & 0.0987  & 0.1420  & \textcolor{blue}{\textbf{0.1534}} & 99\%    & 0.0425  & 0.0423  & 0.0530  & \textcolor{blue}{\textbf{0.0765}} \\
     \midrule
    \textit{Avg.}   & 0.1590  &  0.1461  & 0.1574  & \textcolor{blue}{\textbf{0.1757}} & \textit{Avg.}   & 0.0800  & 0.0809 & 0.0801  & \textcolor{blue}{\textbf{0.0973}} \\
    \midrule
    Computers & \multicolumn{4}{c}{Recall@10} & Photo & \multicolumn{4}{c}{Recall@10} \\
    \midrule
    MR & SVGA  & ITR   & FP  & ARB  & MR & SVGA  & ITR   & FP  & ARB \\
    \midrule

    40\%    & 0.0430  & 0.0430  & 0.0435  & \textcolor{blue}{\textbf{0.0459}} & 40\%    & 0.0440  & 0.0428  & 0.0442  & \textcolor{blue}{\textbf{0.0465}} \\
    50\%   & 0.0430  & 0.0429  & 0.0430  & \textcolor{blue}{\textbf{0.0458}} & 50\%   & 0.0440  & 0.0425  & 0.0441  & \textcolor{blue}{\textbf{0.0459}} \\

    60\%   & 0.0437  & 0.0446  & 0.0425  & \textcolor{blue}{\textbf{0.0449}} & 60\%    & 0.0446  & 0.0434  & 0.0434  & \textcolor{blue}{\textbf{0.0455}} \\

    70\%    & 0.0412  & 0.0420  & 0.0425  & \textcolor{blue}{\textbf{0.0446}} & 70\%    & 0.0433  & 0.0425  & 0.0432  & \textcolor{blue}{\textbf{0.0447}} \\

    80\%   & 0.0410  & 0.0408  & 0.0429  & \textcolor{blue}{\textbf{0.0450}} & 80\%   & 0.0427  & 0.0416  & 0.0430  & \textcolor{blue}{\textbf{0.0445}} \\
    90\%    & 0.0369  &  0.0377 &  0.0418 & \textcolor{blue}{\textbf{0.0434}} & 90\%    & 0.0394  &   0.0382&  0.0423 & \textcolor{blue}{\textbf{0.0437}} \\
    99\%    & 0.0284  & 0.0264  & 0.0410  & \textcolor{blue}{\textbf{0.0420}} & 99\%    & 0.0268  & 0.0278  & 0.0412  & \textcolor{blue}{\textbf{0.0434}} \\
 \midrule
    \textit{Avg.}   &  0.0415 &  0.0396 &  0.0427 & \textcolor{blue}{\textbf{0.0449}} & \textit{Avg.}   & 0.0430  & 0.0418 & 0.0434  & \textcolor{blue}{\textbf{0.0451}} \\
    \bottomrule
    \end{tabular}%
  \label{tabm}%
\end{table}%

\subsection{Training Time Comparison Verification (\textbf{Q6})}
\begin{figure}[h]
\centering
\subfloat[Cora]{
\includegraphics[width=0.22\textwidth]{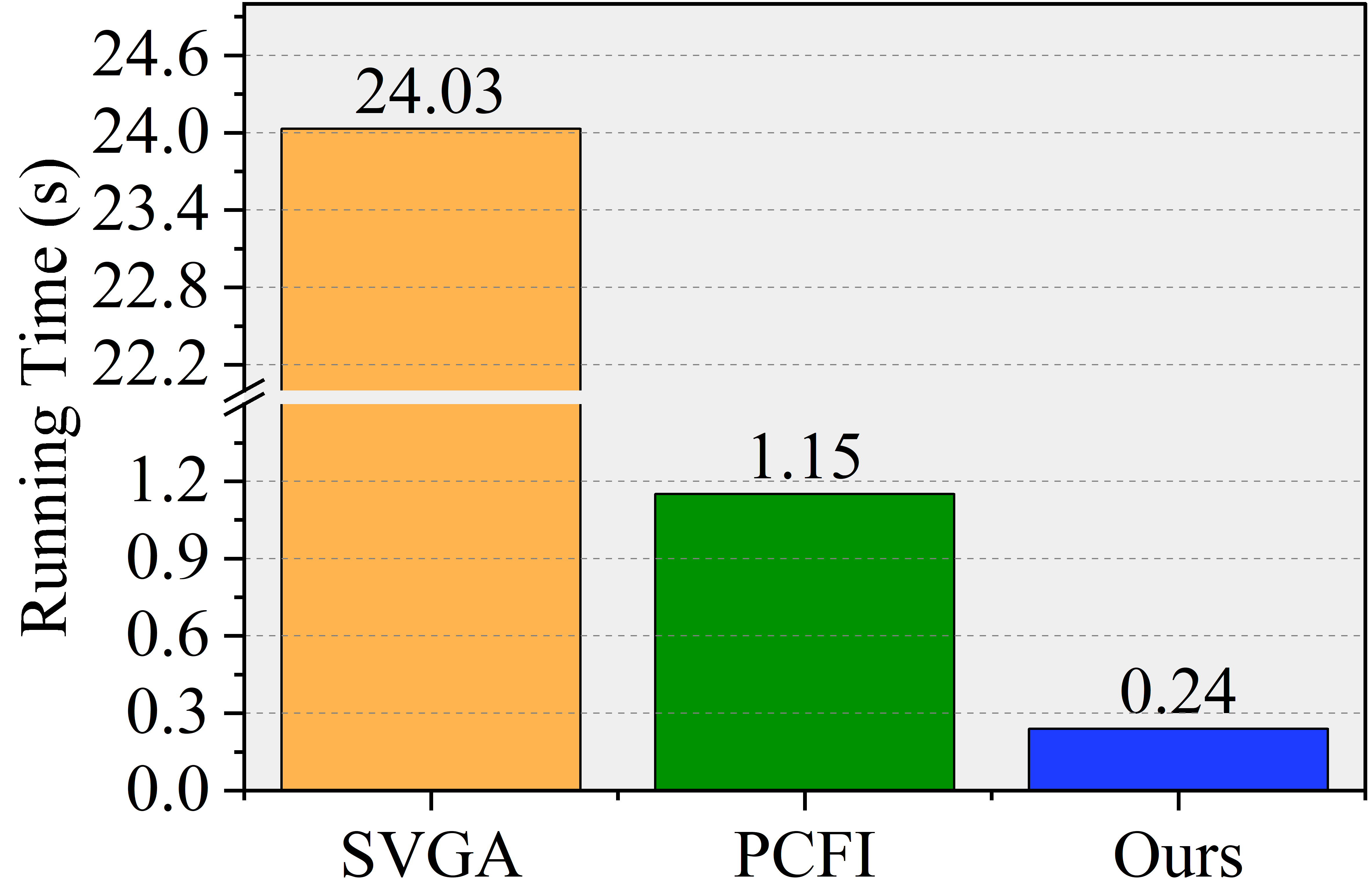}
}
\subfloat[Computers]{
\includegraphics[width=0.22\textwidth]{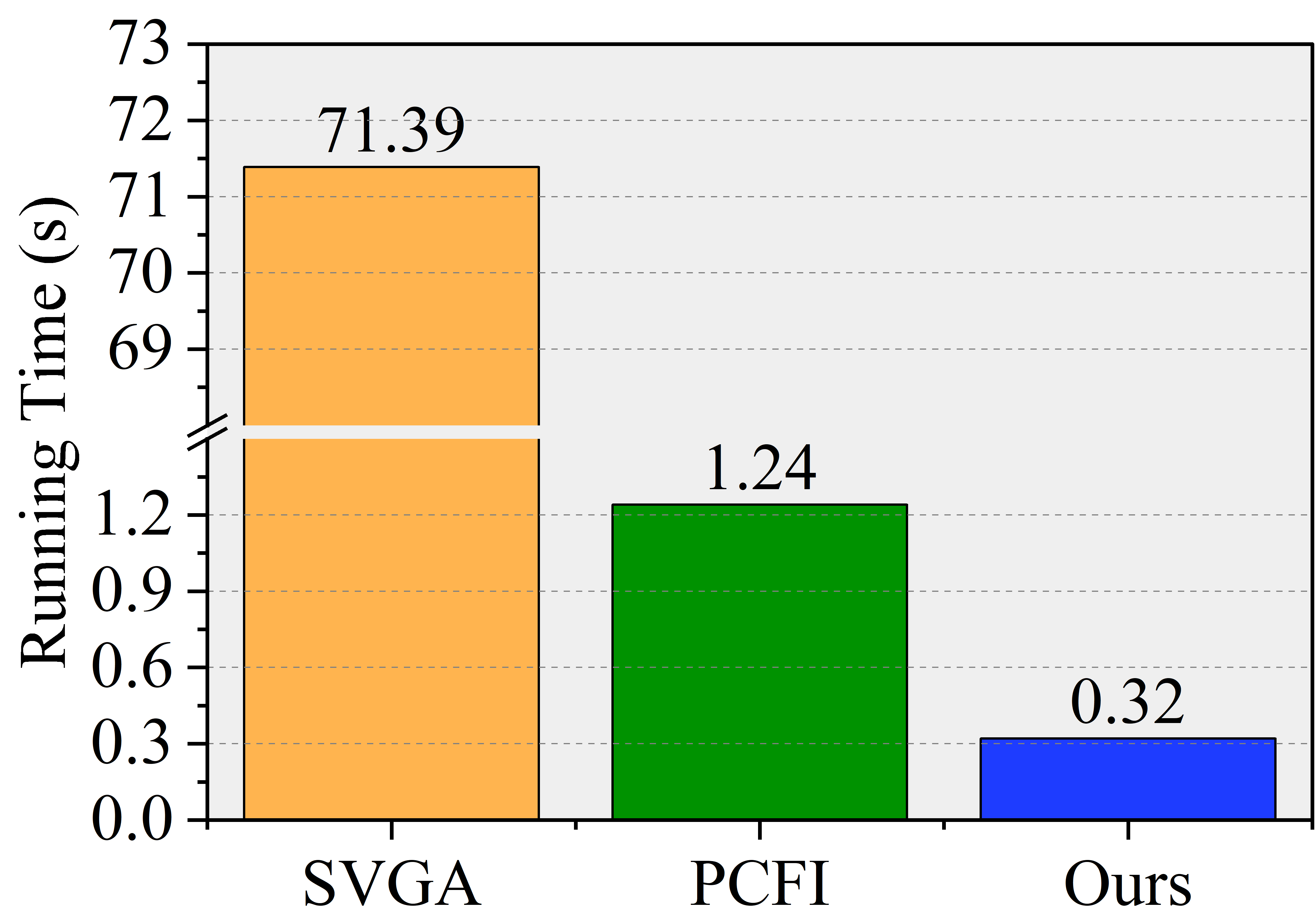}
}
\caption{Comparison of running time.}
\label{fig_3}
\end{figure}

Since propagation methods do not require gradient descent operations, they naturally have an advantage over deep generative methods. Figure \ref{fig_3} shows that, under the same hardware conditions (NVIDIA GeForce RTX 3090 24G) and with SVGA's loss converged, as well as 20 propagation iterations for both PCFI and ARB, ARB significantly outperforms the baseline methods SVGA (deep learning method) and PCFI (propagation method) in terms of training speed. Specifically, ARB is approximately \textbf{163} times faster than SVGA and \textbf{4.4} times faster than PCFI on average.

For propagation methods, Table \ref{taba2} is conducted with the same number of propagation iterations \(l = 20\). It can be observed that our ARB method incurs almost no additional runtime compared to FP. However, PCFI is approximately four times slower than FP. This is primarily because PCFI introduces confidence calculation, which involves searching neighborhood distances and computing the correlation matrix, thereby increasing computational complexity. Additionally, this method exceeded memory capacity on the Ogbn-Products dataset, forcing it to switch to CPU computation, which limited its scalability on large graphs. Overall, the analysis indicates that while ARB has computational efficiency comparable to FP, it significantly outperforms FP in terms of reconstruction accuracy.

Thus, in response to Q6, our conclusion is: \textbf{ARB is highly computationally efficient and low in complexity, making it well-suited for scaling to large graphs.}

\begin{table}[htbp]
  \centering
  \caption{Comparison of running time (s). Best improvement is \textcolor{blue}{\textbf{blue}}.}
      \renewcommand\arraystretch{1.2}
   \renewcommand\tabcolsep{3.5pt}
    \begin{tabular}{ccccc}
    \toprule
    Method & Cora  & CiteSeer & PubMed & Computers \\
    \midrule
    FP    & 0.2327  & 0.2653  & 0.2434  & 0.3152  \\
    PCFI  & 1.1521(×4.95)  & 1.2196(×4.60)  & 1.2136(×4.99)  & 1.2383(×3.93)  \\
    Ours  & 0.2351(\color{blue}{\textbf{×1.01}})  & 0.2654(\color{blue}{\textbf{×1.01}})  & 0.2549(\color{blue}{\textbf{×1.05}})  & 0.3160(\color{blue}{\textbf{×1.01}})  \\
    \midrule
    Method & Photo & CS    & Ogbn-Arxiv & Ogbn-Products \\
    \midrule
    FP    & 0.2548  & 0.8377  & 0.3659  & 16.2063  \\
    PCFI  & 1.1614(×4.56)  & 2.1147(×2.52)  & 1.3760(×3.76)  & OOM. \\
    Ours  & 0.2560(\color{blue}{\textbf{×1.01}})  & 0.8850(\color{blue}{\textbf{×1.06}})  & 0.3700(\color{blue}{\textbf{×1.01}})  & 16.5091(\color{blue}{\textbf{×1.02}})  \\
    \bottomrule
    \end{tabular}%
  \label{taba2}%
\end{table}%

\section{Conclusion}\label{s6}
This paper presents \textit{AttriReBoost} (ARB), a novel method for reconstructing missing attributes in graph data through a propagation-based approach. ARB introduces two key innovations: redefining boundary conditions and incorporating virtual edges, which are specifically designed to address the cold start problem in attribute-missing graphs. The method operates without relying on gradient-based learning, offering a simplified and computationally efficient solution. Theoretical analysis rigorously proves ARB's convergence, and empirical evaluations demonstrate its superior performance in attribute reconstruction and downstream node classification, with a notable reduction in training time. ARB's efficiency and scalability position it as a competitive solution in the field of graph-based learning. 

Future research will focus on three main directions. First, we will explore the relationship between missing attributes and propagation dynamics, aiming to understand how attribute sparsity impacts information flow and design more robust propagation mechanisms. Second, we will investigate alternative methods for constructing virtual edges, such as using graph generative models, structural similarities, or domain-specific heuristics. Finally, we plan to integrate ARB with GNNs as a pre-filling processor, developing adaptive mechanisms to handle varying levels of attribute incompleteness and diverse graph structures, expanding ARB's applicability to a wider range of real-world scenarios.


\bibliographystyle{IEEEtran}
\bibliography{IEEEabrv,tcyb}

\begin{IEEEbiography}[{\includegraphics[width=1in,height=1.1in,clip,keepaspectratio]{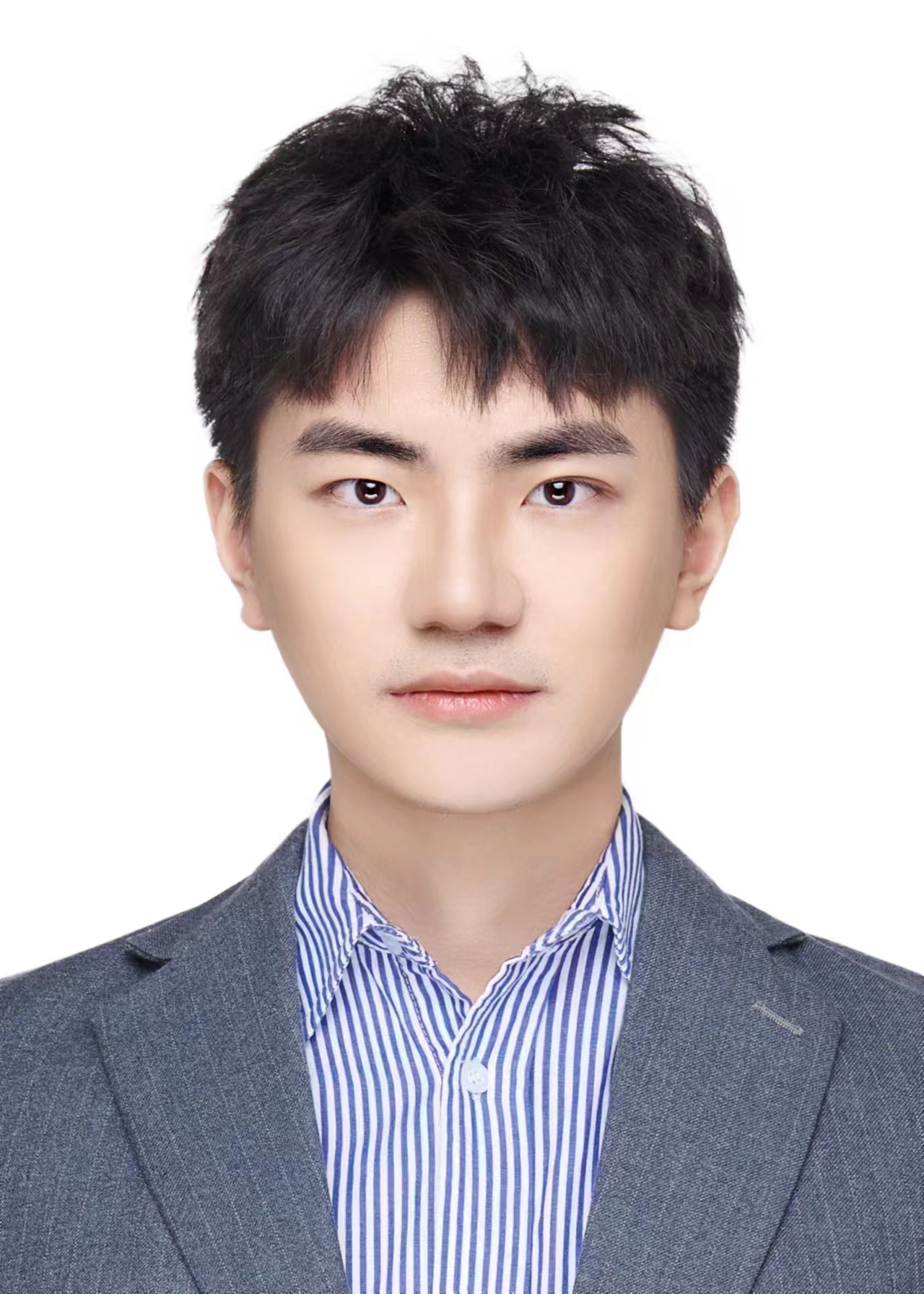}}]
{Mengran Li} is currently pursuing a Ph.D. degree at the Guangdong Key Laboratory of Intelligent Transportation Systems, School of Intelligent Systems Engineering, Sun Yat-sen University, Shenzhen, P.R. China. He received an M.S. degree in Control Science and Engineering from the Beijing Key Laboratory of Multimedia and Intelligent Software Technology, Beijing University of Technology, Beijing, P.R. China, in 2023. His research interests include graph neural networks, data mining, and complex network optimization.
\end{IEEEbiography}

\begin{IEEEbiography}[{\includegraphics[width=1in,height=1.1in,clip,keepaspectratio]{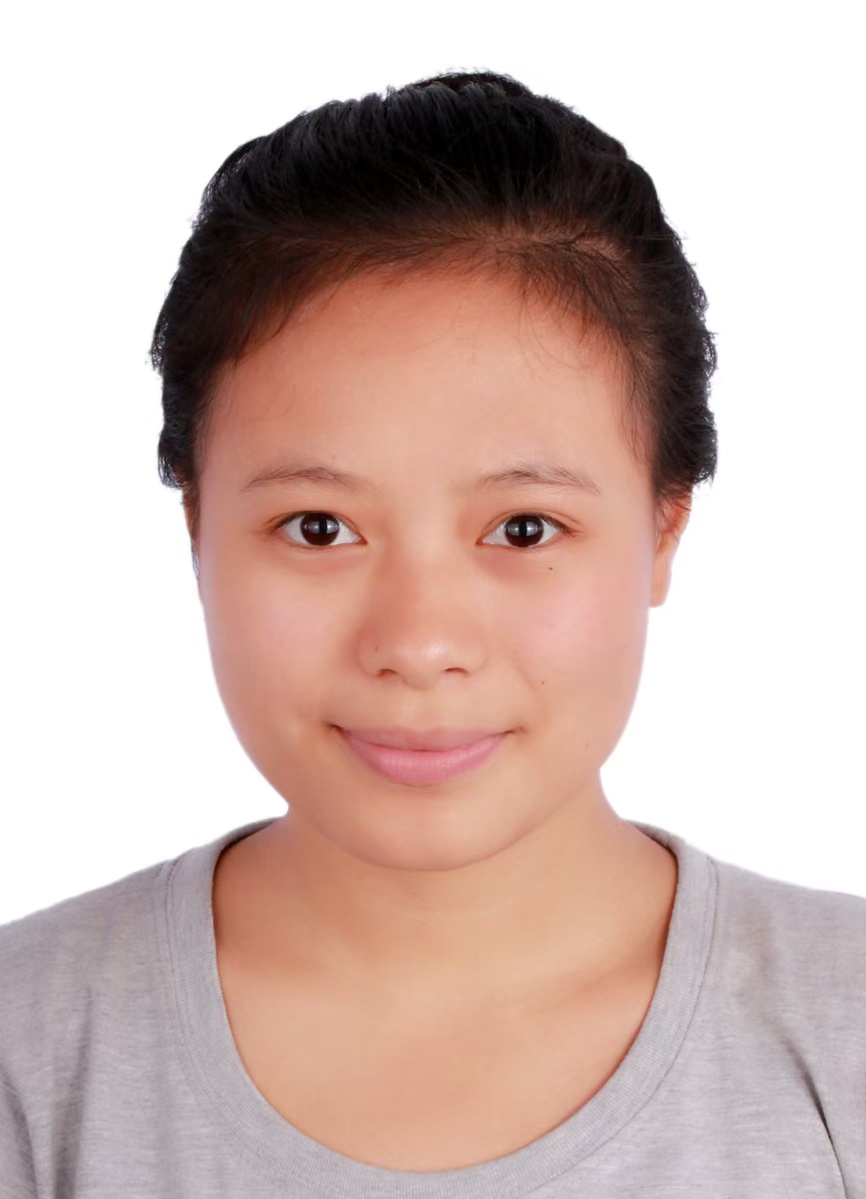}}]
{Chaojun Ding} is currently working as a Research Assistant at the Guangdong Key Laboratory of Intelligent Transportation Systems, School of Intelligent Systems Engineering, Shenzhen Campus of Sun Yat-sen University, Shenzhen, P.R. China. She received her master's degree in Applied Mathematics from P.R. China University of Geosciences. Her research interests include data mining and distributed graph computing.
\end{IEEEbiography}

\begin{IEEEbiography}[{\includegraphics[width=1in,height=1.25in,clip,keepaspectratio]{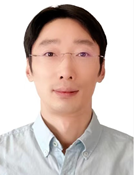}}]
{Junzhou Chen} received his Ph.D. in Computer Science and Engineering from the Chinese University of Hong Kong in 2008. Between March 2009 and February 2019, he served as a Lecturer and later as an Associate Professor at the School of Information Science and Technology at Southwest Jiaotong University. He is currently an Associate Professor at School of Intelligent Systems Engineering at Shenzhen Campus of Sun Yat-sen University. His research interests include computer vision, machine learning and intelligent transportation systems.
\end{IEEEbiography}

\begin{IEEEbiography}[{\includegraphics[width=1in,height=1.25in,clip,keepaspectratio]{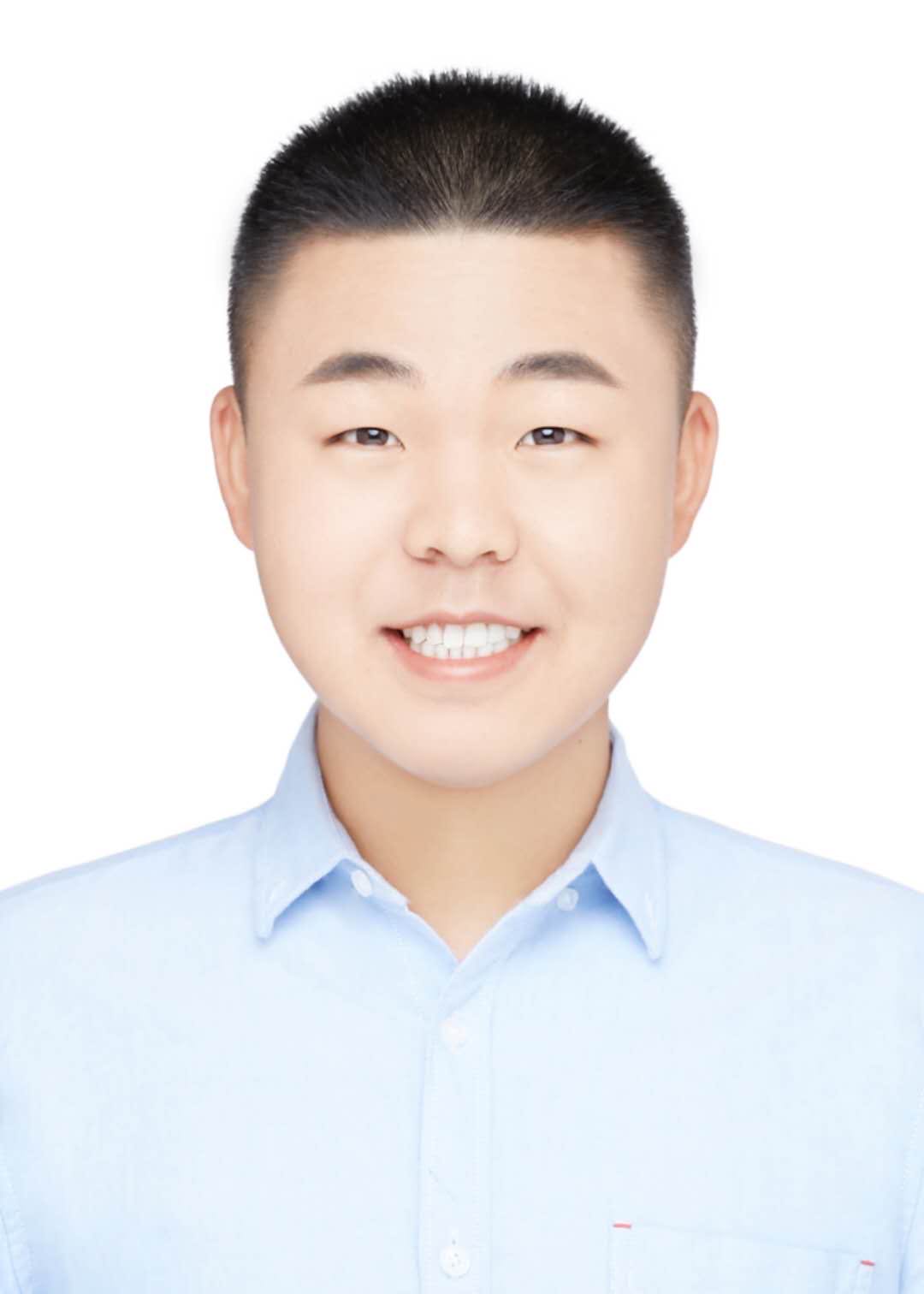}}]
{Wenbin Xing} received his B.S. degree in Computer Science and Technology from Wuhan University of Technology, Wuhan, P.R. China, in 2024. He is pursuing his master's degree at Sun Yat-sen University in Shenzhen, P.R. China. His current research interests include autonomous driving, traffic big data, deep learning, and spatial-temporal modeling.
\end{IEEEbiography}

\begin{IEEEbiography}[{\includegraphics[width=1in,height=1.25in,clip,keepaspectratio]{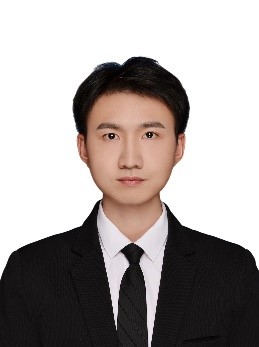}}]
{Cong Ye} received the B.Sc. degree in vehicle engineering from Hefei University of Technology, Hefei, P.R. China, in 2020, and the M.S. degree in vehicle engineering from Jilin University, Changchun, P.R. China, in 2023. He is currently working toward the Ph.D. degree in Electronic Information at the School of Intelligent Engineering, Sun Yat-sen University in Shenzhen, P.R. China. His research interests include vehicle dynamics, x-by-wire chassis, and autonomous driving.
\end{IEEEbiography}

\begin{IEEEbiography}[{\includegraphics[width=1in,height=1.25in,clip,keepaspectratio]{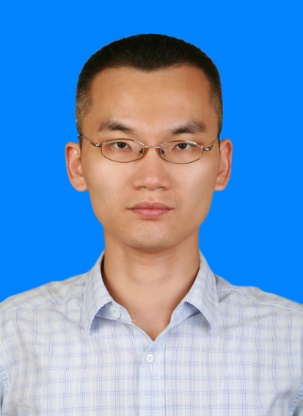}}]
{Ronghui Zhang} received a B.Sc. (Eng.) from the Department of Automation Science and Electrical Engineering, Hebei University, Baoding, P.R. China, in 2003, an M.S. degree in Vehicle Application Engineering from Jilin University, Changchun, P.R. China, in 2006, and a Ph.D. (Eng.) in Mechanical \& Electrical Engineering from Changchun Institute of Optics, Fine Mechanics and Physics, the Chinese Academy of Sciences, Changchun, P.R. China, in 2009. After finishing his post-doctoral research work at INRIA, Paris, France, in February 2011, he is currently an Associate Professor with Guangdong Key Laboratory of Intelligent Transportation System, School of Intelligent Systems Engineering, Shenzhen Campus of Sun Yat-sen University, Shenzhen, Guangdong, P.R. China. His current research interests include computer vision, intelligent control, and ITS. 
\end{IEEEbiography}

\begin{IEEEbiography}[{\includegraphics[width=1in,height=1.25in,clip,keepaspectratio]{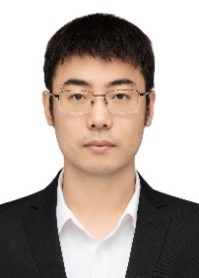}}]
{Songlin Zhuang} received the B.E. degree in automation and the Ph.D. degree in control science and engineering from the Harbin Institute of Technology, Harbin, P.R. China, in 2014 and 2019, respectively. From 2019 to 2020, he was a Postdoctoral Fellow with the Department of Mechanical and Industrial Engineering, University of Toronto, Toronto, Canada. From 2020 to 2023, he was a Postdoctoral Fellow with the Department of Mechanical Engineering, University of Victoria, Victoria, Canada. He is currently a Professor with Yongjiang Laboratory, Ningbo, P.R. China. His research interests include micro/nano-manipulation and control theory. He serves as a Technical Editor of IEEE/ASME Transactions on Mechatronics.
\end{IEEEbiography}


\begin{IEEEbiography}[{\includegraphics[width=1in,height=1.25in,clip,keepaspectratio]{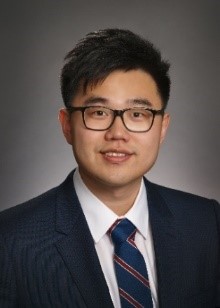}}]
{Jia Hu} works as a ZhongTe Distinguished Chair in Cooperative Automation in the College of Transportation Engineering at Tongji University. Before joining Tongji, he was a research associate at the Federal Highway Administration, USA (FHWA). He serves as an associate editor for IEEE Trans. Intell. Transp. Syst., IEEE Trans. Intell. Veh., ASCE J. Transp. Eng., and IEEE Open J. Intell. Transp. Syst..
He is also an assistant editor of the J. Intell. Transp. Syst., an advisory editorial board member for Transp. Res. Part C, an associate editor for the IEEE Intell. Veh. Symp. since 2018, and an associate editor for the IEEE Intell. Transp. Syst. Conf. since 2019.
\end{IEEEbiography}


\begin{IEEEbiography}[{\includegraphics[width=1in,height=1.25in,clip,keepaspectratio]{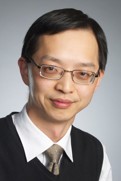}}]
{Tony Z. Qiu} is a Professor in the Faculty of Engineering at the University of Alberta, Canada Research Chair Professor in Cooperative Transportation Systems, and Director of the Centre for Smart Transportation. 
Dr. Tony Qiu received his PhD degree from the University of Wisconsin-Madison, and worked as a Post-Doctoral Researcher in the California PATH Program at the University of California, Berkeley, before joining the University of Alberta.
Dr. Tony Qiu has been awarded the Minister’s Award of Excellence in 2013, the Faculty of Engineering Annual Research Award in 2015-2016, and the ITS Canada Annual Innovation and R\&D Award in 2016 and 2017. His research interest includes traffic operation and control, traffic flow theory, and traffic model analytics. He has published more than 180 papers in international journals and academic conferences, and has 7 awarded patents and 5 pending application patents.
\end{IEEEbiography}

\begin{IEEEbiography}[{\includegraphics[width=1.2in,height=1.25in,clip,keepaspectratio]{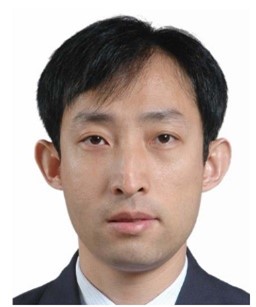}}]
{Huijun Gao} (Fellow, IEEE) received the Ph.D. degree in control science and engineering from the Harbin Institute of Technology, Harbin, P.R. China, in 2005.
Since 2004, he has been with the Harbin Institute of Technology, where he is currently the Chair Professor and the Director of the Research Institute of Intelligent Control and Systems. From 2005 to 2007, he carried out his post-doctoral research with the Department of Electrical and Computer Engineering, University of Alberta, Edmonton, AB, Canada. His research interests include intelligent and robust control, robotics, mechatronics, and their engineering applications.
Dr. Gao is a member of Academia Europaea and the Vice President of the IEEE Industrial Electronics Society. He was a recipient of the 2022 Dr.-Ing. Eugene Mittelmann Achievement Award and the 2023 Norbert Wiener Award. He is a Distinguished Lecturer of the IEEE Systems, Man and Cybernetics Society. He is/was the Editor-in-Chief of IEEE/ASME Transactions on Mechatronics, the Co-Editor-in-Chief of IEEE Transactions on Industrial Electronics, and an Associate Editor of Automatica, IEEE Transactions on Cybernetics, and IEEE Transactions on Industrial Informatics.

\end{IEEEbiography}

\newpage



\newpage
{\appendices

\section*{Supplementary Appendix}
\subsection{Proof of New Boundary Conditions}
\paragraph*{Statement 1} presents the optimization loss for Redefinition of Boundary Conditions to dynamically adjust the initialization of known nodes.

Define:
\begin{equation}
\mathbf I_k^0 = \text{diag}(\{\lambda_1,\lambda_2,\cdots,\lambda_N\}),
\lambda_i = \begin{cases}
1 &\text{ if } i\in\mathcal{V}_k \\
0 &\text{otherwise}
\end{cases}
\end{equation}

\begin{equation}
\mathbf I_u^0 = \text{diag}(\{\lambda_1,\lambda_2,\cdots,\lambda_N\}),
\lambda_i = \begin{cases}
0 &\text{ if } i\in\mathcal{V}_k \\
1 &\text{otherwise}
\end{cases}
\end{equation}

Then:
\begin{equation}
\begin{aligned}
& \nabla\mathcal{L}(\mathbf X) = \mathbf{LX} + \eta(\mathbf X_k-\mathbf Z_k) = 0 \\
\implies& \mathbf{LX} + \eta \mathbf I_k^0(\mathbf X - \mathbf Z) = 0 \\
\implies& (\mathbf I+\eta \mathbf I_k^0)X = \widetilde {\mathbf A}\mathbf X+\eta \mathbf I_k^0\mathbf Z \\
\implies& ((1+\eta) \mathbf I_k^0 + \mathbf I_u^0)X = \widetilde {\mathbf A}\mathbf X+\eta \mathbf I_k^0 \mathbf Z \\
\implies& \mathbf X = (\frac{1}{1+\eta}\mathbf I_k^0+\mathbf I_u^0) (\widetilde {\mathbf A}\mathbf X+\eta \mathbf I_k^0 \mathbf Z) \\
\implies& \mathbf X = \mathbf I_u^0 \widetilde {\mathbf A}\mathbf X + \mathbf I_k^0(\frac{1}{1+\eta} \widetilde {\mathbf A}\mathbf X+\frac{\eta}{1+\eta}\mathbf Z) \\
\implies& \mathbf X = \mathbf I_u^0 \widetilde {\mathbf A}\mathbf X + \mathbf I_k^0(\beta \widetilde {\mathbf A}\mathbf X+(1-\beta)\mathbf Z) \\
\implies& \begin{bmatrix}\mathbf X_k \\ \mathbf X_u \end{bmatrix} = \begin{bmatrix} \beta (\widetilde {\mathbf A}\mathbf X)_k + (1-\beta)\mathbf Z_k \\ (\widetilde {\mathbf A}\mathbf X)_u \end{bmatrix} \\
\implies& \mathbf X = \beta \widetilde {\mathbf A}\mathbf X + (1-\beta) \begin{bmatrix}\mathbf Z_k \\ (\widetilde {\mathbf A}\mathbf X)_u \end{bmatrix}
\end{aligned}
\end{equation}

So:
\begin{equation}\label{e24}
\begin{cases}
\mathbf X = \widetilde {\mathbf A}\mathbf X, \\
\mathbf X_k = \beta \mathbf X_k + (1-\beta)\mathbf Z_k
\end{cases}
\end{equation}
$\square$ \textit{Convergence Proof}

The Equation \eqref{e24} can be written as:

\begin{equation}
\begin{bmatrix} \mathbf X_k \\ \mathbf X_u \end{bmatrix} = \begin{bmatrix}\beta \mathbf A_{kk} & \beta \mathbf A_{ku} \\ \mathbf A_{uk} & \mathbf A_{uu} \end{bmatrix} \begin{bmatrix} \mathbf X_k \\ \mathbf X_u \end{bmatrix} + \begin{bmatrix} (1-\beta)\mathbf Z_k \\ 0 \end{bmatrix}
\end{equation}

Let
\begin{equation}
K = \begin{bmatrix}\beta \mathbf A_{kk} & \beta \mathbf A_{ku} \\ \mathbf A_{uk} & \mathbf A_{uu} \end{bmatrix}, \quad \mathbf C = \begin{bmatrix} (1-\beta)\mathbf Z_k \\ 0 \end{bmatrix}
\end{equation}

Here $\beta \in [0,1)$, then $0 \le \mathbf K \le \widetilde {\mathbf A}$ elementwise and $\mathbf K \ne \widetilde {\mathbf A}$. According to Section \ref{s3} D in this paper, $\rho(\mathbf K) < 1$, Equation \eqref{e17} converges. This completes the proof. $\blacksquare$

\subsection{Proof of Virtual Edges}
\paragraph*{Statement 1} presents the optimization loss for Virtual Edges, to ensure robust connectivity in the attribute-missing graph.

Define:
\\
\begin{equation}
\mathbf I_k^0 = \text{diag}(\{\lambda_1,\lambda_2,\cdots,\lambda_N\}),
\lambda_i = \begin{cases}
1 &\text{ if } i\in\mathcal{V}_k \\
0 &\text{otherwise}
\end{cases}
\end{equation}
\\
Then:

\begin{equation}
\begin{aligned}
&\nabla \mathcal{L}(\mathbf X) = 0 \\
\implies& \mathbf I_k^0 (\mathbf {LX} + \theta \mathbf L_1 \mathbf X) = 0 \\
\implies& \mathbf I_k^0\left[(1+\theta)\mathbf X-(\widetilde {\mathbf A}+\theta \widetilde {\mathbf A}_1)\mathbf X \right] = 0 \\
\implies& \mathbf I_k^0\left[(1+\theta)\mathbf X-\widetilde {\mathbf A}\mathbf X - \theta (\frac{1}{N-1}\mathbf J_{N\times N}-\frac{1}{N-1}\mathbf I)\mathbf X \right] = 0 \\
\implies& \mathbf I_k^0\left[(1+\theta)\mathbf X - \widetilde {\mathbf A}\mathbf X - \theta(\frac{N}{N-1}\overline {\mathbf X} - \frac{1}{N-1} \mathbf {X}) \right] = 0 \\
\implies& \mathbf I_k^0\left[(1+\theta+\frac{\theta}{N-1})\mathbf X - (\widetilde {\mathbf A}\mathbf X+\theta\frac{N}{N-1}\overline {\mathbf X} ) \right] = 0 \\
\implies& \mathbf I_k^0\left[ \mathbf X - \frac{N-1}{N+\theta N -1}\widetilde {\mathbf A}\mathbf X - \frac{\theta N}{N+\theta N -1}\overline {\mathbf X} \right] = 0\\
\implies& \mathbf I_k^0\left[ \mathbf X - \alpha \widetilde {\mathbf A}\mathbf X - (1-\alpha) \overline {\mathbf X} \right] = 0
\end{aligned}
\end{equation}

So:

\begin{equation}\label{e17}
\begin{cases}
\mathbf X = \alpha \widetilde {\mathbf A}\mathbf X + (1-\alpha)\overline {\mathbf X} \\
\mathbf X_k = \mathbf Z_k
\end{cases}
\end{equation}

$\square$ \textit{Convergence Proof}

Let:

\begin{equation}
\mathbf B = \alpha \widetilde {\mathbf A} + (1-\alpha)\frac{1}{N}\mathbf J
\end{equation}
where $\mathbf J$ is an all-ones matrix, $\rho(\frac{1}{N}\mathbf J)=1$. So $\mathbf B$ is a strongly connected matrix.
According to $\rho(\widetilde {\mathbf A})\le 1$, $\rho(\mathbf B)\le 1$.
Equation \eqref{e17} can be written as:

\begin{equation}
\mathbf X = \begin{bmatrix} 0 & 0 \\ \mathbf B_{uk} & \mathbf B_{uu} \end{bmatrix}\mathbf X + \begin{bmatrix} \mathbf Z_k \\ 0 \end{bmatrix}
\end{equation}

Let:

\begin{equation}
\mathbf K = \begin{bmatrix} 0 & 0 \\ \mathbf B_{uk} & \mathbf B_{uu} \end{bmatrix} \quad \mathbf C = \begin{bmatrix} \mathbf Z_k \\ 0 \end{bmatrix}
\end{equation}

Then $0 \le \mathbf K \le \mathbf B$ elementwise and $\mathbf K \ne \mathbf B$. According to Section \ref{s3} D in this paper, $\rho(\mathbf K) < 1$, Equation \eqref{e17} converges. This completes the proof. $\blacksquare$

From the above proofs, our ARB can be simplified into two forms, represented by Equations \eqref{e17} and \eqref{e24}, each tailored to different graph structures. The hyperparameters $\alpha$ and $\beta$ are not set based on intuition but are derived through the optimization equation. Moreover, when both are set to 1, ARB degenerates into FP.

}
\vfill

\end{document}